\documentclass{article}

\newcommand{\attn}{3D full attention}
\newcommand{\Attn}{3D Full Attention}
\newcommand{\SpatialHead}{Spatial Head}
\newcommand{\spatialhead}{spatial head}
\newcommand{\TemporalHead}{Temporal Head}
\newcommand{\temporalhead}{temporal head}

\newcommand{\onlinesample}{online profiling strategy}
\newcommand{\Onlinesample}{Online profiling strategy}
\newcommand{\reorder}{layout transformation}
\newcommand{\sys}{SVG}

\NewDocumentCommand{\jt}{ mO{} }{\textcolor{blue}{\textsuperscript{\textit{JT}}\textsf{\small[#1]}}}

\usepackage{microtype}
\usepackage{graphicx}
\usepackage{subfigure}
\usepackage{booktabs} 
\usepackage{url}
\usepackage{graphicx} 
\usepackage{amsmath}
\usepackage{algorithm}
\usepackage{algpseudocode}
\usepackage{subfigure}
\usepackage{booktabs}
\usepackage{caption}
\usepackage{multirow}
\usepackage{arydshln}
\usepackage{xcolor}
\usepackage{tcolorbox} 
\usepackage{colortbl}
\usepackage{makecell}
\usepackage{bigstrut}
\usepackage{pifont}
\usepackage{listings}
\usepackage{enumitem}

\definecolor{darkgreen}{rgb}{0.15, 0.75, 0.15}
\definecolor{cvprblue}{rgb}{0.21,0.49,0.74}
\definecolor{lightblue}{rgb}{0.90, 0.95, 0.99}

\usepackage{hyperref}

\usepackage[arxiv]{icml2025}

\usepackage{amsmath}
\usepackage{amssymb}
\usepackage{mathtools}
\usepackage{amsthm}

\usepackage[capitalize,noabbrev]{cleveref}

\theoremstyle{plain}

\theoremstyle{definition}

\theoremstyle{remark}

\usepackage[textsize=tiny]{todonotes}

\icmltitlerunning{Sparse VideoGen: Accelerating Video Diffusion Transformers with Spatial-Temporal Sparsity}

\begin{document}

\twocolumn[
\icmltitle{Sparse VideoGen: Accelerating Video Diffusion Transformers with Spatial-Temporal Sparsity}



\icmlsetsymbol{equal}{*}

\begin{icmlauthorlist}
\icmlauthor{Haocheng Xi}{equal,ucb}
\icmlauthor{Shuo Yang}{equal,ucb}
\icmlauthor{Yilong Zhao}{ucb}
\icmlauthor{Chenfeng Xu}{ucb}
\icmlauthor{Muyang Li}{mit}
\icmlauthor{Xiuyu Li}{ucb}
\icmlauthor{Yujun Lin}{mit}
\icmlauthor{Han Cai}{nvidia}
\icmlauthor{Jintao Zhang}{tsinghua}
\icmlauthor{Dacheng Li}{ucb}
\icmlauthor{Jianfei Chen}{tsinghua}
\icmlauthor{Ion Stoica}{ucb}
\icmlauthor{Kurt Keutzer}{ucb}
\icmlauthor{Song Han}{mit,nvidia}
\end{icmlauthorlist}

\icmlaffiliation{ucb}{University of California, Berkeley}
\icmlaffiliation{mit}{MIT}
\icmlaffiliation{nvidia}{NVIDIA}
\icmlaffiliation{tsinghua}{Tsinghua University}

\icmlcorrespondingauthor{Chenfeng Xu}{xuchenfeng@berkeley.edu}

\icmlkeywords{Machine Learning, ICML}

\vskip 0.3in
]



\printAffiliationsAndNotice{\icmlEqualContribution} 

\begin{abstract}

Diffusion Transformers (DiTs) dominate video generation but their high computational cost severely limits real-world applicability, usually requiring tens of minutes to generate a few seconds of video even on high-performance GPUs.
This inefficiency primarily arises from the quadratic computational complexity of \attn{} with respect to the context length. In this paper, we propose a training-free framework termed Sparse VideoGen (\textbf{\sys{}}) that leverages the inherent sparsity in \attn{} to boost inference efficiency. We reveal that the attention heads can be dynamically classified into two groups depending on distinct sparse patterns: (1) \textit{\SpatialHead{}}, where only spatially-related tokens within each frame dominate the attention output, and (2) \textit{\TemporalHead{}}, where only temporally-related tokens across different frames dominate. Based on this insight, \sys{} proposes an \onlinesample{} to capture the dynamic sparse patterns and predicts the type of attention head. Combined with a novel hardware-efficient tensor \reorder{} and customized kernel implementations, \sys{} achieves up to $2.28\times$ end-to-end speedup on CogVideoX-v1.5, $2.33\times$ on HunyuanVideo, and $1.51×\times$ on Wan 2.1, while preserving generation quality. Our code is open-sourced and is available at \href{https://github.com/svg-project/Sparse-VideoGen}{https://github.com/svg-project/Sparse-VideoGen}.



\end{abstract}

\begin{figure}[h!]
    \centering
    \includegraphics[width=0.95\linewidth]{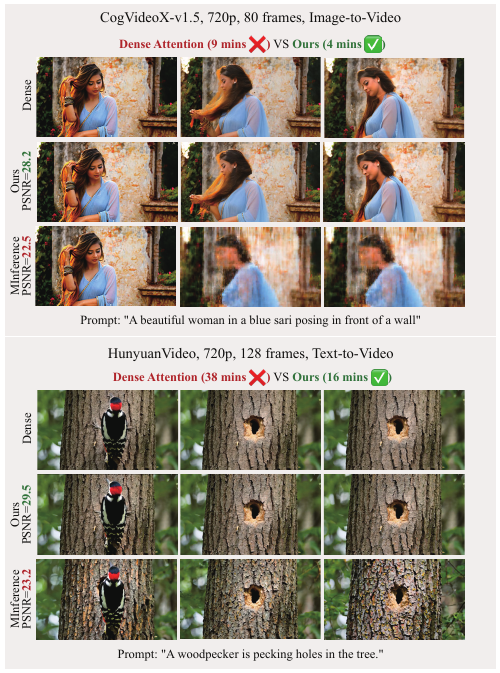}
    \caption{\textbf{\sys{} accelerates video generation while maintaining high quality}. On CogVideoX-v1.5-I2V and Hunyuan-T2V, our method achieves a $2.28\times$ and $2.33\times$ speedup with high PSNR. In contrast, MInference \citep{jiang2024minference} fails to maintain pixel fidelity (significant blurring in the first example) and temporal coherence (inconsistencies in the tree trunk in the second example).}
    \label{fig:teaser}
\end{figure}

\section{Introduction}\label{sec:introduction}

Diffusion Transformers (DiTs)~\citep{peebles2023scalable} have recently emerged as a transformative paradigm for generative tasks, achieving state-of-the-art results in image generation. This success has been naturally carried over to video generation, with models adapting from a spatial 2D attention to a spatiotemporal \attn{}~\cite{arnab2021vivit,yang2024cogvideox,kong2024hunyuanvideo}, resulting in high-fidelity and temporally consistent outputs.
Close-sourced models such as Sora and Kling, and open-sourced models including Wan 2.1~\citep{wang2025wan21}, CogVideo~\citep{hongcogvideo}, and HunyuanVideo~\citep{kong2024hunyuanvideo}, have showcased impressive capabilities in applications ranging from animation~\cite{guoanimatediff,feng2024explorative} to physical world simulation~\citep{liu2023world}. 

\begin{figure}[t]
    \centering
    \begin{minipage}[t]{\linewidth}
        \includegraphics[width=\linewidth]{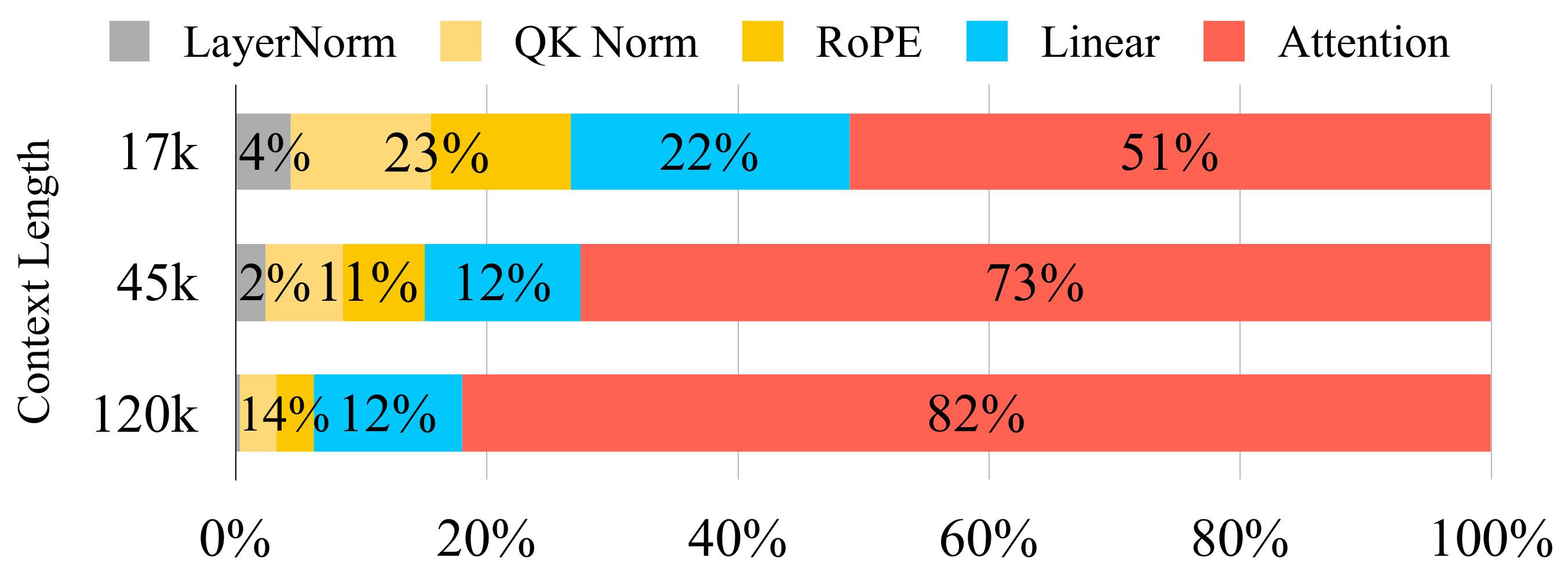}
        \caption{\textbf{Attention dominates the computation in video diffusion models.} For CogVideoX-v1 and -v1.5 with $17$k and $45$k context length, attention takes $51$\% and $73$\% of the latency, respectively. For HunyuanVideo with $120$k context length, attention can take over $80$\% amount of the runtime latency.}
        \label{fig:intro-op-breakdown}
    \end{minipage}
\end{figure}

Despite significant advances in generating high-quality videos, the deployment of video generation models remains challenging due to their substantial computation usage. For instance, HunyuanVideo requires almost an hour on an NVIDIA A100 GPU to generate only a $5$-second video, where the \attn{} accounts for more than $80$\% of end-to-end runtime (Figure~\ref{fig:intro-op-breakdown}). Moreover, due to the \textit{quadratic computational complexity} with respect to the context length~\cite{dao2022flashattentionfastmemoryefficientexact}, the attention can be much more dominant as the resolution and number of frames increase, as shown in Figure~\ref{fig:intro-op-breakdown}. 


Fortunately, attention in transformers is well-known for its sparsity, offering a great opportunity to reduce redundant computation. For example, in large language models (LLMs),  a small portion of the tokens can dominate the attention output~\cite{zhang2023h2oheavyhitteroracleefficient, xiao2024efficientstreaminglanguagemodels, tang2024questqueryawaresparsityefficient}. Therefore, the computation can be dramatically reduced by only computing the attention among such important tokens, while still maintaining generation accuracy. However, existing methods cannot be directly applied to DiTs (as shown in Table~\ref{table:accuracy_efficiency_benchmark}), as video data has fundamentally different sparsity patterns from text data.

\begin{figure*}[ht]
    \centering
    \includegraphics[width=\textwidth]{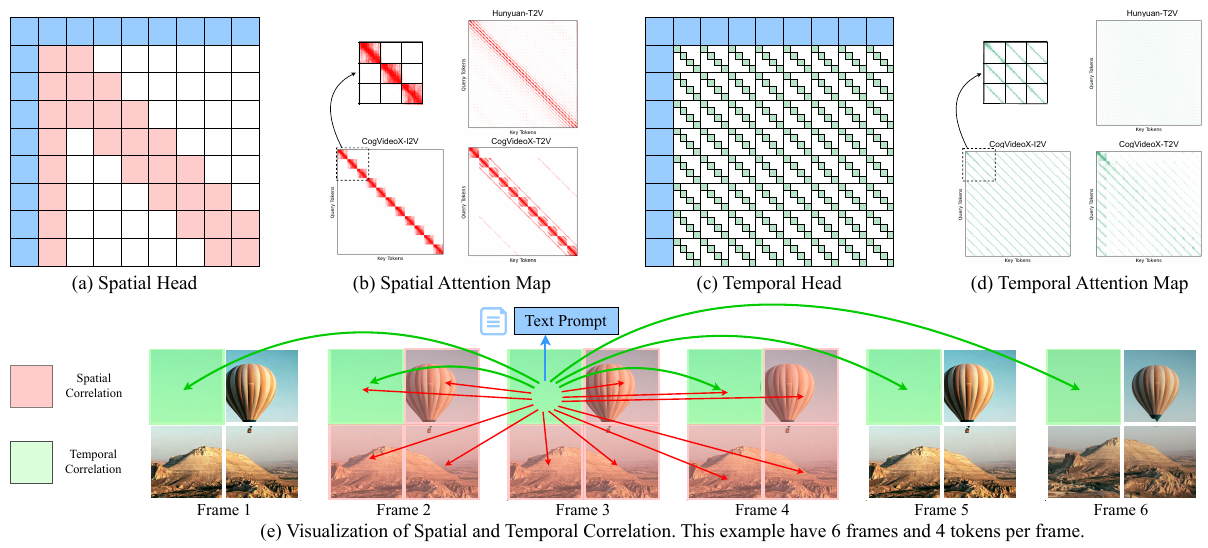} 
        \captionof{figure}{We observe two types of attention maps with distinct sparse patterns: \textit{spatial map (b)} and \textit{temporal map (d)}. Based on the attention map, we classify all attention heads into \textit{\SpatialHead{} (a)} and \textit{\TemporalHead{} (c)}, which contribute to the spatial and temporal consistency of generated videos respectively. As visualized in (e), \spatialhead{} primarily focuses on all tokens within the same frame (painted as red). In contrast, \temporalhead{} attends to tokens at the same position across all frames (painted as green).
        }
        \label{fig:spatial-temporal-illustration} 
\end{figure*}

Our key observation is that attention heads in DiTs exhibit inherent sparsity in two categories: \textit{\SpatialHead{}} and \textit{\TemporalHead{}}, based on their distinct sparsity patterns. As shown in Figure~\ref{fig:spatial-temporal-illustration}, \spatialhead{} mainly focuses on tokens that reside within the same frame, which determines the spatial structures of generated videos. In contrast, \temporalhead{} attends to tokens at the same spatial location across all frames, contributing to the temporal consistency. Therefore, the computation for both types of heads can be greatly reduced by only calculating the attended tokens.

Despite the theoretical speedup, leveraging sparsity for end-to-end acceleration is still challenging. Firstly, sparsity patterns are highly dynamic across different denoising steps and input prompts. It necessitates an online method to identify sparsity patterns without incurring overhead. Secondly, some sparsity patterns are unfriendly to hardware accelerators. For example, \temporalhead{} computes over noncontiguous data that cannot be fed to GPU's tensor cores, resulting in significant efficiency degradations~\cite{ye2023sparsetircomposableabstractionssparse}.

To tackle these challenges, we propose Sparse VideoGen (\sys{}), a training-free framework that accelerates video DiTs with the following novel designs: (1) To efficiently identify the best sparsity pattern for each attention head, \sys{} introduces an \onlinesample{} with minimal overhead ($\sim$3\%). It randomly samples 1\% tokens from each attention head and processes sampled tokens with full attention computation and two distinct sparse attentions (\spatialhead{} and \temporalhead{}). Finally, the sparse pattern with a lower error compared to the full attention is selected for each head. (2) To improve hardware efficiency, \sys{} proposes a novel \reorder{}, which reorders the noncontiguous sparsity pattern of \temporalhead{} into a compact and hardware-friendly sparsity pattern.

We prototype \sys{} with customized kernel implementation by Triton~\cite{Tillet2019TritonAI} and FlashInfer~\cite{ye2025flashinferefficientcustomizableattention} and evaluate \sys{}'s accuracy and efficiency on representative open video generative models including CogVideoX-v1.5-I2V, CogVideoX-v1.5-T2V, and HunyuanVideo-T2V. \sys{} delivers significant efficiency improvements, achieving an end-to-end speedup of up to $2.33\times$ while maintaining high visual quality with a PSNR of up to 29, outperforming all prior methods. Additionally, we show that \sys{} is compatible with FP8 quantization, enabling additional efficiency gains without compromising quality. We summarize our key contributions as follows:

\begin{itemize}[leftmargin=*]
    \vspace{-5pt}
    \item 
    We conduct in-depth analysis of video DiTs' sparse patterns, revealing two inherent sparse attention patterns (\temporalhead{} and \spatialhead{}) for efficient video generation. 
    \item We propose \sys{}, a training-free sparse attention framework comprising an efficient \onlinesample{} and an efficient inference system for accurate and efficient video generation.
    \item \sys{} delivers significant speedup while maintaining good video generation quality, paving the way for practical applications of video generative models. 
\end{itemize}

\begin{figure*}[t]
    \begin{minipage}{\textwidth}
        \centering
        \includegraphics[width=0.95\textwidth]{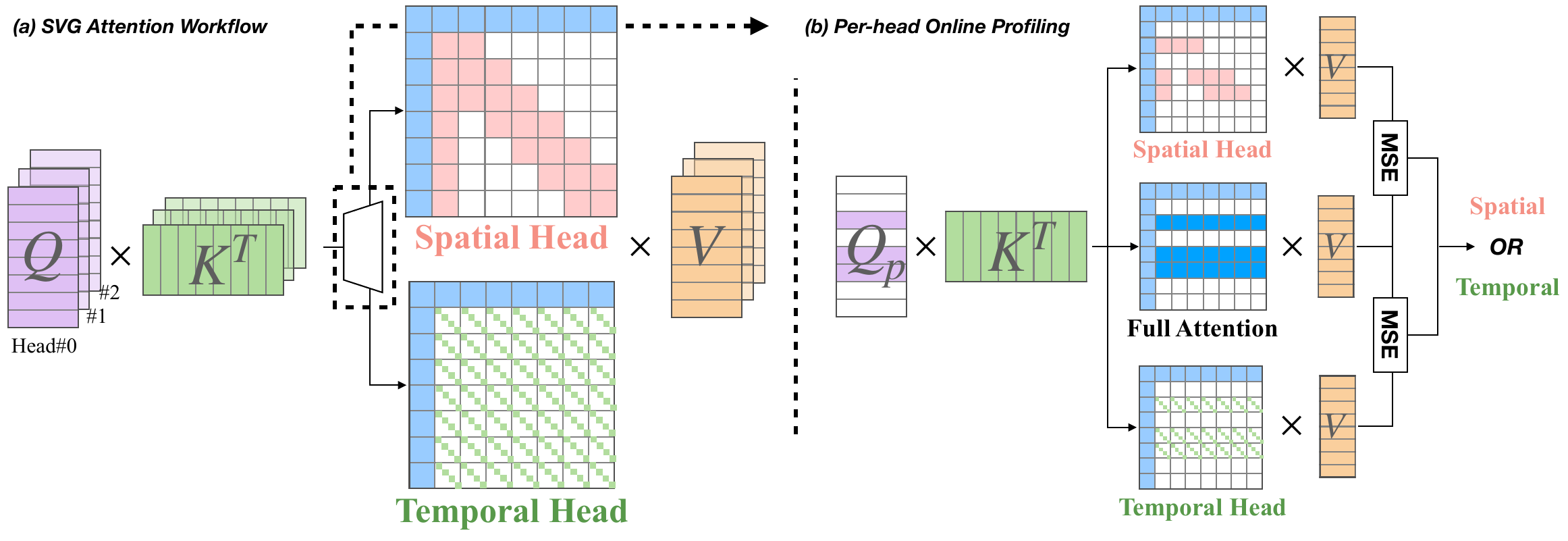}
        \captionof{figure}{Overview of \sys{} framework. (a) During generation, \sys{} adaptively classifies each attention head as either a \textit{\spatialhead{}} or a \textit{\temporalhead{}} and applies a dedicated sparse attention computation accordingly. (b) This adaptive classification is driven by \onlinesample{}, which extracts a small portion of $Q$, denoted as $Q_p$, to perform both spatial and temporal attention computations. \sys{} then selects the attention patter that yields the minimal MSE compared to full attention, ensuring accurate classification.}
        \label{fig:algorithm_overview}
    \end{minipage}
\end{figure*}

\section{Related Work}
\label{sec:related_works}

\subsection{Efficient diffusion models}

\noindent\textbf{Decreasing the denoising steps.}
Most diffusion models employ SDEs that require many sampling steps \citep{song2019generative, ho2020denoising, meng2022sdedit}. To address this, DDIM \citep{song2020denoising} approximates them with an ODE; subsequent techniques refine ODE paths and solvers \citep{lu2022dpm, lu2022dpm++, liu2022flow, liu2023instaflow} or incorporate consistency losses \citep{song2023consistency, luo2023latent}. Distillation-based methods \citep{yin2024improved, yin2024one} train simpler, few-step models. However, these require expensive re-training or fine-tuning—impractical for most video use cases. In contrast, our approach directly uses off-the-shelf pre-trained models without any additional training.

\noindent\textbf{Diffusion model compression.}
Weight compression through quantization is a common tactic \citep{li2023q, zhao2024vidit, li2024svdquant}, pushing attention modules to INT8 \citep{zhang2025sageattention} or even INT4/FP8 \citep{zhang2024sageattention2}. Other work proposes efficient architectures \citep{xie2024sana,cai2024condition,chen2025pixart} or high-compression autoencoders \citep{chen2024deep} to improve performance. Our Sparse VideoGen is orthogonal to these techniques and can incorporate them for additional gains.

\noindent\textbf{Efficient system implementation.}
System-level optimizations focus on dynamic batching \citep{kodaira2023streamdiffusion, liang2024looking}, caching \citep{chen2024delta, zhao2024pab}, or hybrid strategies \citep{lv2024fastercache, liu2024timestep}. While these methods can improve throughput, their output quality often drops below a PSNR of 22. By contrast, our method preserves a PSNR above 30, thus substantially outperforming previous approaches in maintaining output fidelity.

\subsection{Efficient attention methods}\label{subsec:efficient_attention}


\looseness=-1
\noindent\textbf{Sparse attention in LLMs.}
Recent research on sparse attention in language models reveals diverse patterns to reduce computational overhead. StreamingLLM \cite{xiao2023efficient} and LM-Infinite \cite{han2023lm} observe that attention scores often concentrate on the first few or local tokens, highlighting temporal locality. H2O \cite{zhang2023h2o}, Scissorhands \cite{liu2024scissorhands} and DoubleSparsity \cite{yang2024posttrainingsparseattentiondouble} identify a small set of ``heavy hitter'' tokens dominating overall attention scores. TidalDecode \cite{yang2024tidaldecode} shows that attention patterns across layers are highly correlated, while DuoAttention \cite{xiao2024duoattention} and MInference \cite{jiang2024minference} demonstrate distinct sparse patterns across different attention heads. However, these methods focus on token-level sparsity and do not leverage the inherent redundancy of video data.

\noindent\textbf{Linear and low-bit attention.}
Another direction involves linear attention \cite{cai2023efficientvit,xie2024sana,wang2020linformer,choromanski2020rethinking,yu2022metaformer,katharopoulos2020transformers}, which lowers complexity from quadratic to linear, and low-bit attention \cite{zhang2025sageattention,zhang2024sageattention2}, which operates in reduced precision to accelerate attention module. Sparse VideoGen is orthogonal to both approaches: it can be combined with techniques like FP8 attention while still benefiting from the video-specific spatial and temporal sparsity in video diffusion models.

\section{Motivation and Analysis}
\label{sec:sparse_patterns}

\subsection{\Attn{} shows instinct sparsity}
\label{subsec:sparse-pattern}
We identify that \attn{} possess inherent sparsity, characterized by different distinct sparse patterns tailored for different functions~\citep{xiao2024duoattention}. We deeply investigate the sparsity nature across various text-to-video and image-to-video models and identify two types of attention heads based on sparse patterns: \textit{\SpatialHead{}} and \textit{\TemporalHead{}}, as shown in Figure.~\ref{fig:spatial-temporal-illustration}.


\looseness=-1
\textbf{\SpatialHead{}.} As illustrated in Figure~\ref{fig:spatial-temporal-illustration}(a-b), \spatialhead{} primarily focuses its attention scores on spatially-local tokens. This leads to the attention map exhibiting a \textit{block-wise layout}. Since pixels within the same frame are tokenized into contiguous tokens, \spatialhead{} attends exclusively to pixels within the same frame and its adjacent frames. This property is essential for maintaining the spatial consistency of generated videos. In \spatialhead{}, the block size relates to the \textit{number of tokens per frame}. 

\looseness=-1
\textbf{\TemporalHead{}.} In contrast, \temporalhead{} exhibits a \textit{slash-wise layout} with a constant interval (Figure~\ref{fig:spatial-temporal-illustration}(c-d)). Since each frame is tokenized into a fixed number of tokens $L$, pixels occupying the same spatial position across different frames are arranged at a stride of $L$. Consequently, \temporalhead{} captures information from the token with the same spatial position across multiple frames. This pattern is important for ensuring temporal consistency in video generation\footnote{We hypothesize that this occurs because the majority of the training data consists of slow-motion videos, making the temporal head's focus on tokens with the same spatial position in several frames adequate to maintain temporal consistency.}.

In addition to the spatial and temporal patterns, we observe that the text prompts and the first frame hold significant attention scores for both spatial and \temporalhead{}, which aligns with previous investigations~\cite{xiao2024efficientstreaminglanguagemodels,shen2024longvuspatiotemporaladaptivecompression,su2025akvqvlattentionawarekvcache}. Therefore, we include these tokens in both the spatial and temporal head.

\subsection{Sparse attention achieves lossless accuracy}
\label{subsec:sparse-accurate}
Furthermore, we find that directly applying sparse patterns to corresponding heads does not hurt the quality of generated videos.
We demonstrate this by evaluating CogVideoX-v1.5 and HunyuanVideo on VBench~\citep{huang2023vbenchcomprehensivebenchmarksuite} with sparse attention. We determine the sparse pattern by computing full attention along with two different sparse mechanisms (\spatialhead{} and \temporalhead{}) for each attention head and denoising step. The sparse pattern with the lowest mean squared error (MSE) relative to full attention is chosen for further inference. This approach achieves a PSNR over 29, showing that the right sparse pattern maintains the high quality of generated videos.

However, despite its accuracy, this strategy does not provide practical efficiency benefits, as full attention computation is still required to determine the optimal sparse pattern. We will address this issue in Sec~\ref{subsec:sampling_based_pattern_selection}.

\subsection{Sparse attention promises theoretical speedup}
\label{sec:sparse-theoretical-speedup}
\looseness=-1
Instead of computing full attention, sparse attention selectively processes only the important tokens based on sparsity patterns, leading to significant computational savings. We analyze the theoretical computation saving below.

Given a model configuration of hidden dimension $H$, number of tokens per frame $L$, and number of total frames $N$, the total computation (FLOPS) for each full attention is $2\cdot2\cdot(LN)^2\cdot H=4L^2N^2H$. For \spatialhead{}, assuming each head only attends to nearby $c_s$ frames, the computation is reduced to $(2\cdot2\cdot L^2H)\cdot c_sN$, resulting in a sparsity of $\frac{c_s}{N}$. For \temporalhead{}, assuming each token only attends $c_t$ tokens across all the frames, the computation is $(2\cdot2\cdot N^2 H)\cdot c_tL$, with a sparsity of $\frac{c_t}{L}$. 
Since both $c_s$ and $c_t$ are typically much smaller compared to $N$ and $L$ respectively, the sparsity can easily achieve $30$\%.
E.g., the aforementioned CogVideoX-v1.5-T2V achieves a sparsity of $31$\% for both spatial and \temporalhead{} while maintaining an average of $29.99$ PSNR.

Despite the theoretical speedup, the temporal head can not be directly translated into real speedup since the pattern is hardware-inefficient. We will discuss our practical solution in Sec.~\ref{subsec:frame_token_rearrangement} and prove it can achieve theoretical speedup in Sec.~\ref{subsec:kernel_level_efficiency}. Note that we do not include the text prompts and first frame in the theoretical calculation for simplicity, as they are constant and small compared to the remaining part.

\section{Methodology}
\label{sec:methodology}

\begin{algorithm}[t]
\caption{Online Profiling Strategy}
\label{alg:sample_based_pattern_selection}
\definecolor{codeblue}{rgb}{0.25,0.65,0.5}
\definecolor{codeblue2}{rgb}{0,0,1}
\lstset{
  backgroundcolor=\color{white},
  basicstyle=\fontsize{7.2pt}{7.2pt}\ttfamily\selectfont,
  columns=fixed,
  breaklines=true,
  captionpos=b,
  commentstyle=\fontsize{7.2pt}{7.2pt}\color{codeblue},
  keywordstyle=\fontsize{7.2pt}{7.2pt}\color{codeblue2},
    emph={mask},
    emphstyle={\color[RGB]{255,52,179}},
}
\begin{lstlisting}[language=python]
# Q, K, V, O: [B, H, S, D] - query, key, value, output
# S:              - Total Token Number E.g., 18k
# t:              - Sampled Token Number. E.g., 32

# Sample the Indices
indices = sample_indices(S, t) # (t,)
Q_i = Q[:, :, indices, :]

# Get the attention masks
mask_spatial = gen_spatial_mask()[:, :, indices, :]
mask_temporal = gen_temporal_mask()[:, :, indices, :]

# Compute sampled attention score
# Shape: [B, H, t, D]
O_full = mask_attention(Q_i, K, V, None)
O_spatial = mask_attention(Q_i, K, V, mask_spatial)
O_temporal = mask_attention(Q_i, K, V, mask_temporal)

# Calculate MSE and get best mask
# Shape: [B, H]
MSE_s = (O_full - O_spatial).norm().mean(dim=(2,3))
MSE_t = (O_full - O_temporal).norm().mean(dim=(2,3))
best_mask_config = (MSE_s < MSE_t)

\end{lstlisting}
\end{algorithm}

In this section, we introduce \sys{}, a training-free framework designed to exploit the sparse patterns of \attn{} while addressing practical deployment challenges through careful design. To identify sparse patterns, \sys{} employs an \onlinesample{} (Sec~\ref{subsec:sampling_based_pattern_selection}). To effectively utilize sparsity, \sys{} introduces a hardware-efficient \reorder{}, which enables real-world hardware acceleration (Sec~\ref{subsec:frame_token_rearrangement}). Additionally, by integrating techniques such as customized kernels and quantization (Sec~\ref{subsec:other-optimization}), \sys{} significantly accelerates video generation without compromising generation quality.

\subsection{\Onlinesample{} for sparsity identification}
\label{subsec:sampling_based_pattern_selection}

As discussed in Sec~\ref{subsec:sparse-pattern}, all attention heads can be classified and sparsified into \spatialhead{} and \temporalhead{}. However, we find that such sparse patterns can be \textit{highly dynamic} across different denoising steps and input data. E.g., a certain head can be a \spatialhead{} for one prompt while being a \temporalhead{} given another. This dynamic nature necessitates an efficient online sparsity identification method, which classifies attention heads on the fly without extra overhead.

To this end, \sys{} proposes an \textit{\onlinesample{}}. Instead of computing the entire full attention to identify sparse attention, \sys{} only samples a subset of input rows ($x$\%) and calculates results with both the spatial and temporal sparsity patterns. By choosing the one with the lower MSE compared to full attention, \sys{} can efficiently approximate the oracle identification method discussed in Sec~\ref{subsec:sparse-accurate}. We detail the profiling process in Algorithm~\ref{alg:sample_based_pattern_selection}.

To demonstrate the effectiveness of the proposed method, we conduct a sensitivity test on profiling ratio $x$ with CogVideoX-v1.5-I2V. As shown in Table~\ref{table:sensitivity-sampling}, profiling only $1$\% can achieve up to $31.1$ PSNR, with only $3$\% runtime overhead compared to full attention.

\begin{table*}[!t]
\centering
\caption{Quality and efficiency benchmarking results of \sys{} and other baselines. 
}
\label{table:accuracy_efficiency_benchmark}
\resizebox{0.95\linewidth}{!}{%
\begin{tabular}{c|l|ccccc|ccc}
\toprule
\textbf{Type} & \textbf{Method} & \multicolumn{5}{c|}{\textbf{Quality}} & \multicolumn{3}{c}{\textbf{Efficiency}} \\
\cmidrule(lr){3-7} \cmidrule(lr){8-10}
& & PSNR $\uparrow$ & SSIM $\uparrow$ & LPIPS $\downarrow$ & ImageQual $\uparrow$ & SubConsist $\uparrow$ & FLOPS $\downarrow$ & Latency $\downarrow$ & Speedup $\uparrow$ \\
\midrule
\textbf{I2V} & CogVideoX-v1.5 (720p, 10s, 80 frames) & - & - & - & 70.09\% & 95.37\% & 147.87 PFLOPs &  528s & 1x \\
\midrule
& DiTFastAttn (Spatial-only) & 24.591 & 0.836 & 0.167 & 70.44\% & 95.29\% & 78.86 PFLOPs &   338s  & 1.56x \\
& Temporal-only & 23.839 & 0.844 & 0.157 & 70.37\% & 95.13\% & 70.27 PFLOPs &  327s & 1.61x \\
& MInference & 22.489 & 0.743 & 0.264 & 58.85\% & 87.38\% & 84.89 PFLOPs  & 357s & 1.48x \\
& PAB & 23.234 & 0.842 & 0.145 & 69.18\% & 95.42\% & 105.88 PFLOPs  & 374s & 1.41x \\
\rowcolor{lightblue}
& Ours & \textbf{\textcolor{darkgreen}{28.165}} & \textbf{\textcolor{darkgreen}{0.915}} & \textbf{\textcolor{darkgreen}{0.104}} & 70.41\% & 95.29\% & 74.57 PFLOPs &  237s & \textbf{\textcolor{darkgreen}{2.23x}} \\
\midrule
\textbf{T2V} & CogVideoX-v1.5 (720p, 10s, 80 frames) & - & - & - & 62.42\% & 98.66\% & 147.87 PFLOPs &  528s & 1x \\
\midrule
& DiTFastAttn (Spatial-only) & 23.202 & 0.741 & 0.256 & 62.22\% & 96.95\% & 78.86 PFLOPs &  338s & 1.56x \\
& Temporal-only & 23.804 & 0.811 & 0.198 & 62.12\% & 98.53\% & 70.27 PFLOPs &  327s & 1.61x \\
& MInference & 22.451 & 0.691 & 0.304 & 54.87\% & 91.52\% & 84.89 PFLOPs & 357s  & 1.48x \\
& PAB & 22.486 & 0.740 & 0.234 & 57.32\% & 98.76\% & 105.88 PFLOPs & 374s  & 1.41x \\
\rowcolor{lightblue}
& Ours & \textbf{\textcolor{darkgreen}{29.989}} & \textbf{\textcolor{darkgreen}{0.910}} & \textbf{\textcolor{darkgreen}{0.112}} & 63.01\% & 98.67\% & 74.57 PFLOPs &   232s & \textbf{\textcolor{darkgreen}{2.28x}} \\
\midrule
\textbf{T2V} & HunyuanVideo (720p, 5.33s, 128 frames) & - & - & - & 66.11\% & 93.69\% & 612.37 PFLOPs &  2253s & 1x \\
\midrule
& DiTFastAttn (Spatial-only) & 21.416 & 0.646 & 0.331 & 67.33\% & 90.10\% & 260.48 PFLOPs &   1238s & 1.82x \\
& Temporal-only & 25.851 & 0.857 & 0.175 & 62.12\% & 98.53\% & 259.10 PFLOPs &  1231s & 1.83x \\
& MInference & 23.157 & 0.823 & 0.163 & 63.96\% & 91.12\% & 293.87 PFLOPs &  1417s  &  1.59x \\
\rowcolor{lightblue}
& Ours & \textbf{\textcolor{darkgreen}{29.546}} & \textbf{\textcolor{darkgreen}{0.907}} & \textbf{\textcolor{darkgreen}{0.127}} & 65.90\% & 93.51\% & 259.79 PFLOPs &   1171s & 1.92x \\
\rowcolor{lightblue}
& Ours + FP8 & \textbf{\textcolor{darkgreen}{29.452}} & \textbf{\textcolor{darkgreen}{0.906}} & \textbf{\textcolor{darkgreen}{0.128}} & 65.70\% & 93.51\% & 259.79 PFLOPs &  968s & \textbf{\textcolor{darkgreen}{2.33x}} \\
\bottomrule
\end{tabular}%
}
\end{table*}

\subsection{Hardware-efficient \reorder{}}
\label{subsec:frame_token_rearrangement}

\begin{figure}[t]
    \centering
    \includegraphics[width=\columnwidth]{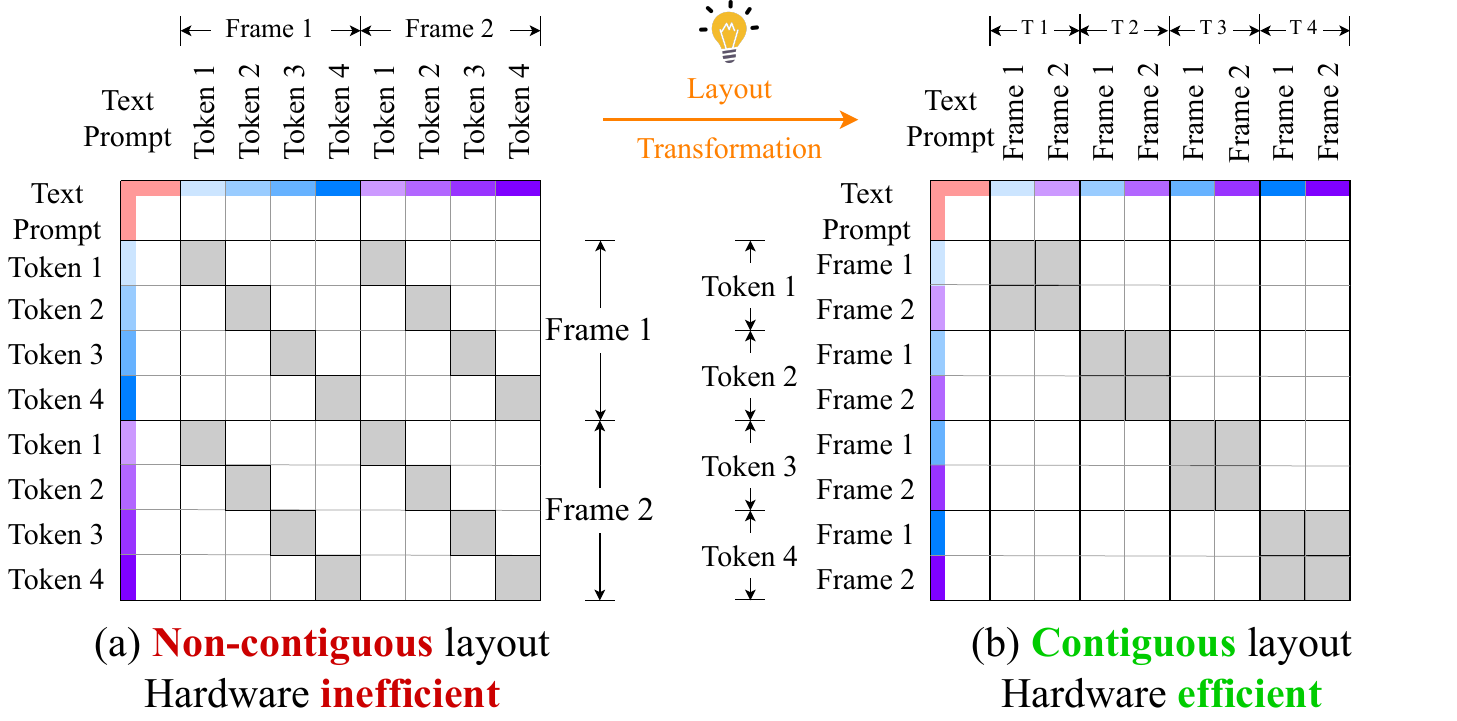} 
    \caption{Visualization of hardware-efficient \reorder{}. (a) Non-contiguous sparsity layout of \temporalhead{}, which is hardware inefficient due to the contiguous layout required by hardware accelerators. (b) Contiguous layout generated by transposing the token-major tensor into a frame-major one, which can be efficiently handled by block sparse attention.}
    \label{fig:frame-token-rearrangement-figure}
\end{figure}

\looseness=-1
Despite the high sparsity in attention computation, speedups are limited without a hardware-efficient sparsity layout~\cite{ye2023sparsetircomposableabstractionssparse,zheng2023pitoptimizationdynamicsparse}. For instance, NVIDIA’s Tensor Core, a matrix-matrix multiplication accelerator, requires at least $16$ contiguous elements along each dimension to use. However, \temporalhead{} exhibits a sparse layout of non-contiguous elements with a stride of $L$ (i.e., number of tokens per frame). This sparsity pattern prevents effective utilization of Tensor Core, thereby constraining overall efficiency.

To tackle this, \sys{} introduces a \textit{\reorder{}} strategy that transforms the sparsity layout of \temporalhead{} into a hardware-efficient one. As illustrated in Figure~\ref{fig:frame-token-rearrangement-figure}, this strategy transposes a token-major tensor into a frame-major one, which makes the tokens across different frames into a contiguous layout. Such transformation maintains a mathematically equivalent output as attention computation is associative~\cite{dao2022flashattentionfastmemoryefficientexact,dao2019butterfly}. We ablate the effectiveness of the proposed method in Sec~\ref{subsec:kernel_level_efficiency}.

\subsection{Other optimizations}
\label{subsec:other-optimization}

\textbf{Efficient kernel customization.} We notice that current implementations of QK-norm and RoPE suffer from performance issues, due to limited parallelism on small head dimensions (e.g., $64$ in CogVideoX-v1.5). Therefore, we customize those operations with CUDA by a sub-warp reduction implementation, providing up to $5\times$ speedup compared to torch implementation (see Table~\ref{table:small-kernel-speedup-comparison}). We also use Triton to implement fused \onlinesample{} and \reorder{} kernels, followed by a block sparse attention kernel with FlashInfer \citep{ye2025flashinferefficientcustomizableattention}.

\textbf{Quantization.} We further incorporate \textbf{FP8 quantization} into sparse attention~\cite{zhang2025sageattention,zhang2024sageattention2,zhao2024atom}, which further boosts up to $1.3\times$ throughput with minimal accuracy drop as shown in Table~\ref{table:accuracy_efficiency_benchmark}. We also customize an attention kernel that supports both FP8 quantization and block sparse computation.

\section{Experiment}
\label{subsec:experiments}

\begin{figure*}[t!]
    \centering
    \includegraphics[width=\textwidth]{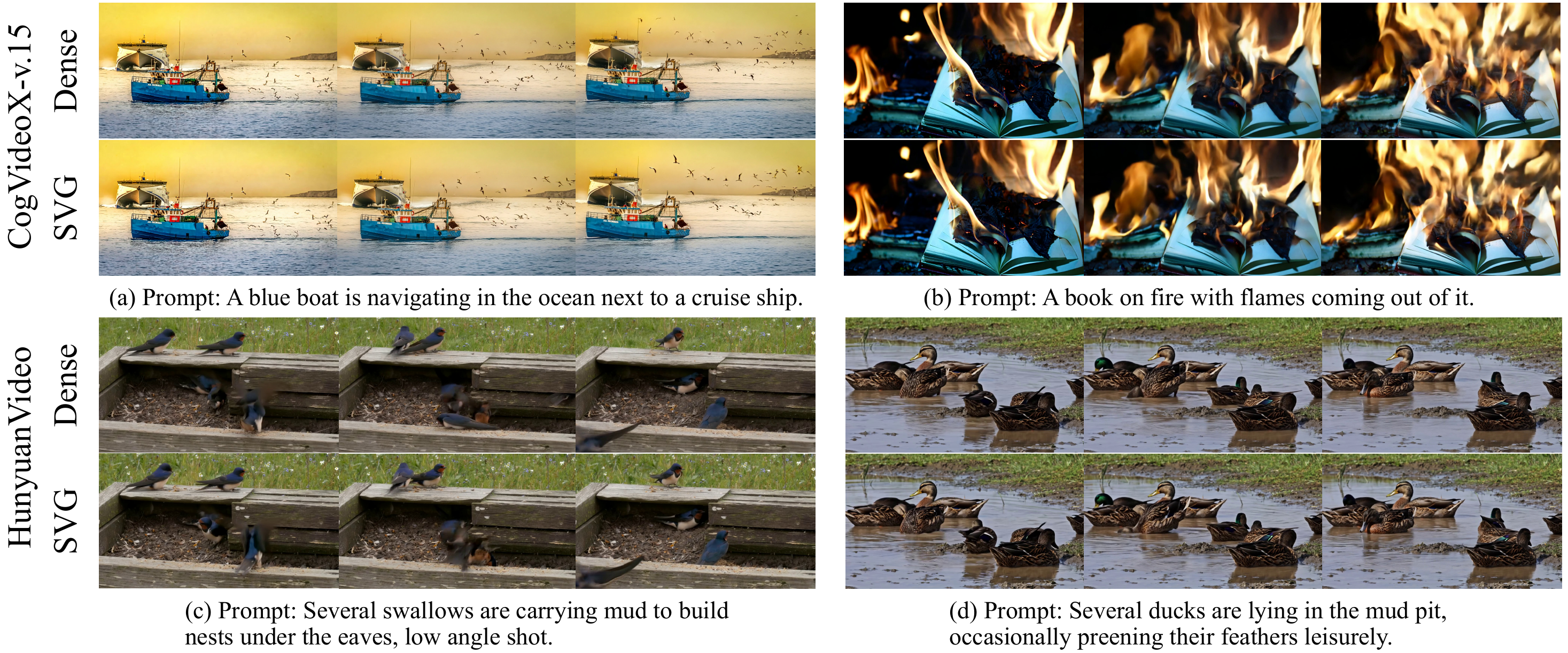} 
        \captionof{figure}{Examples of generated videos by \sys{} and original implementation on CogVideoX-v1.5-I2V and HunyuanVideo-T2V. We showcase four different scenarios: (a) minor scene changes, (b) significant scene changes, (c) rare object interactions, and (d) frequent object interactions. \sys{} produces videos highly consistent with the originals in all cases, maintaining high visual quality.}
        \label{fig:SVG-visualization} 
\end{figure*}

\subsection{Setup}
\label{subsec:experiment_setup}

\textbf{Models.} We evaluate \sys{} on open-sourced state-of-the-art video generation models including CogVideoX-v1.5-I2V, CogVideoX-v1.5-T2V, and HunyuanVideo-T2V to generate $720$p resolution videos. After 3D VAE, CogVideoX-v1.5 consumes $11$ frames with $4080$ tokens per frame in \attn{}, while HunyuanVideo works on $33$ frames with $3600$ tokens per frame.

\textbf{Metrics.} We assess the quality of the generated videos using the following metrics. We use Peak Signal-to-Noise Ratio (PSNR), Learned Perceptual Image Patch Similarity (LPIPS)~\citep{zhang2018perceptual}, Structural Similarity Index Measure (SSIM) to evaluate the generated video's similarity, and use VBench Score~\citep{huang2023vbenchcomprehensivebenchmarksuite} to evaluate the video quality, following common practices in community~\citep{5596999,zhao2024pab,li2024svdquant,li2024distrifusion}. Specifically, we report the imaging quality and subject consistency metrics, represented by ImageQaul and SubConsist in our table.

\textbf{Datasets.} For CogVideoX-v1.5, we generate video using the VBench dataset after prompt optimization, as suggested by CogVideoX~\cite{yang2024cogvideox}. 
For HunyuanVideo, we benchmark our method using the prompt in Penguin Video Benchmark released by HunyuanVideo~\cite{kong2024hunyuanvideo}.


\textbf{Baselines.} We compare \sys{} against sparse attention algorithms DiTFastAttn~\cite{yuan2024ditfastattnattentioncompressiondiffusion} and MInference~\cite{jiang2024minference}. As DiTFastAttn can be considered as \spatialhead{} only algorithm, we also manually implement a \temporalhead{} only baseline named \textit{Temporal-only attention}. We also include a cache-based DiT acceleration algorithm PAB~\cite{zhao2024pab} as a baseline.

\textbf{Parameters.} For MInference and PAB, we use their official configurations. For \sys{}, we choose $c_s$ as $4$ frames and $c_t$ as $1224$ tokens for CogVideoX-v1.5, while $c_s$ as $10$ frames and $c_t$ as $1200$ tokens for HunyuanVideo. Such configurations lead to approximately $30$\% sparsity for both \spatialhead{} and \temporalhead{}, which is enough for lossless generation in general. We skip the first $25$\% denoising steps for all baselines as they are critical to generation quality, following previous works~\cite{zhao2024pab,li2024distrifusion,lv2024fastercache,liu2024timestep}.

\textbf{Visualizations.} We present a comparison of the videos generated by Dense Attention and Sparse VideoGen in Appendix~\ref{appendix: visualization}. Additionally, real video samples are available on Google Drive and can be accessed \href{https://drive.google.com/drive/folders/1jhEpZ69bKfyZWmoy63iS3FhECNnX-AZU?usp=drive_link}{here}.

\begin{figure}[t]
    \centering
    \includegraphics[width=0.95\columnwidth]{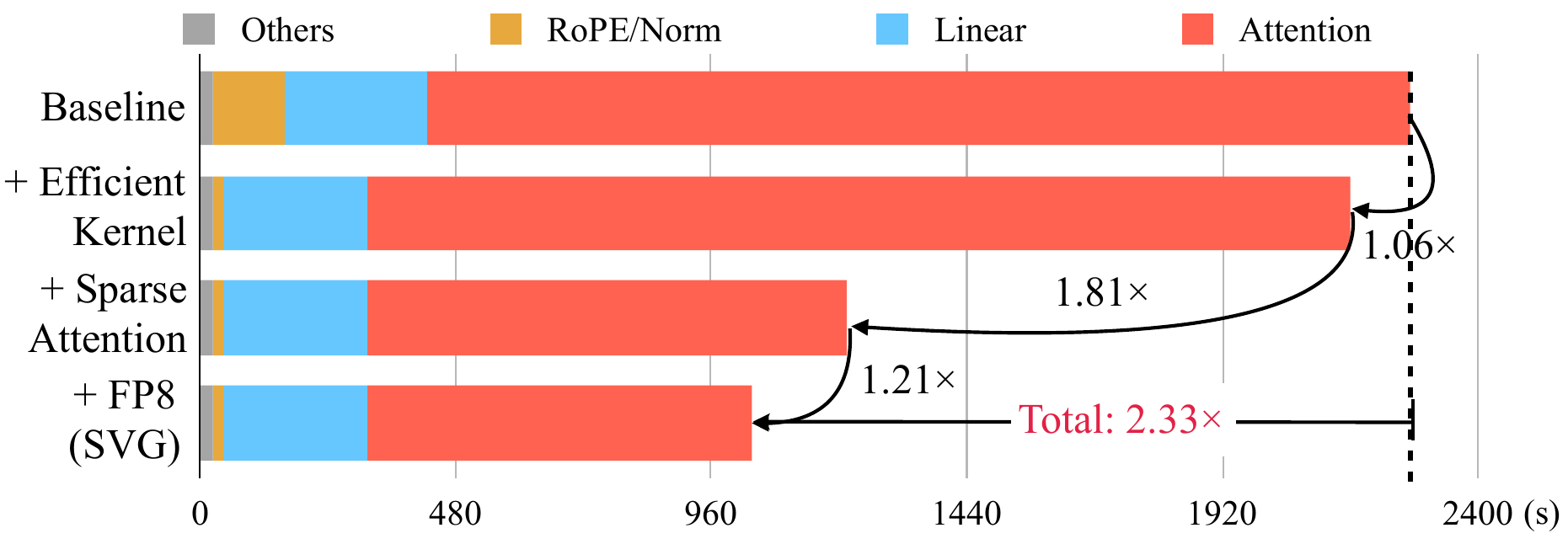} 
    \caption{The breakdown of end-to-end runtime of HunyuanVideo when generating a $5.3$s, $720$p video. \sys{} effectively reduces the end-to-end inference time from $2253$ seconds to $968$ seconds through system-algorithm co-design. Each design point contributes to a considerable improvement, with a total $2.33\times$ speedup.}
    \label{fig:efficiency-breakdown-figure}
\end{figure}

\subsection{Quality evaluation}
\label{subsec:quality_benchmark}
We evaluate the quality of generated videos by \sys{} compared to baselines and report the results in Table~\ref{table:accuracy_efficiency_benchmark}. Results demonstrate that \sys{} \textbf{consistently outperforms} all baseline methods in terms of PSNR, SSIM, and LPIPS while achieving \textbf{higher end-to-end speedup}.

Specifically, \sys{} achieves an average PSNR exceeding \textbf{29.55} on HunyuanVideo and \textbf{29.99} on CogVideoX-v1.5-T2V, highlighting its exceptional ability to maintain high fidelity and accurately reconstruct fine details.
For a visual understanding of the video quality generated by \sys{}, please refer to Figure \ref{fig:SVG-visualization}.

\sys{} maintains both \textbf{spatial and temporal consistency} by adaptively applying different sparse patterns, while all other baselines fail. E.g., since the mean-pooling block sparse cannot effectively select slash-wise temporal sparsity (see Figure~\ref{fig:spatial-temporal-illustration}), MInference fails to account for temporal dependencies, leading to a substantial PSNR drop. Besides, PAB skips computation of \attn{} by reusing results from prior layers, which greatly hurts the quality.

Moreover, \sys{} is compatible with \textbf{FP8 attention quantization}, incurring only a $0.1$ PSNR drop on HunyuanVideo. Such quantization greatly boosts the efficiency by $1.3\times$. Note that we do not apply FP8 attention quantization on CogVideoX-v1.5, as its head dimension of $64$ limits the arithmetic intensity, offering no on-GPU speedups.

\subsection{Efficiency evaluation}
\label{subsec:efficiency_benchmark}

To demonstrate the feasibility of \sys{}, we prototype the entire framework with dedicated CUDA kernels based on FlashAttention~\cite{dao2022flashattentionfastmemoryefficientexact}, FlashInfer~\cite{ye2025flashinferefficientcustomizableattention}, and Triton~\cite{Tillet2019TritonAI}. We first showcase the end-to-end speedup of \sys{} compared to baselines on an H100-80GB-HBM3 with CUDA 12.4. Besides, we also conduct a kernel-level efficiency evaluation. Note that all baselines adopt FlashAttention-2~\cite{dao2022flashattentionfastmemoryefficientexact}.

\begin{table}[t]
\small
\centering
\caption{Inference speedup of customized QK-norm and RoPE compared to PyTorch implementation with different number of frames. We use the same configuration of CogVideoX-v1.5, i.e. $4080$ tokens per frame, $96$ attention heads.}
\label{table:small-kernel-speedup-comparison}
\begin{tabular}{c|cccc}
\toprule
Frame Number & 8 & 9 & 10 & 11  \\
\midrule
QK-norm & 7.44× & 7.45× & 7.46× & 7.47×  \\
\midrule
RoPE & 14.50× & 15.23× & 15.93× & 16.47×   \\
\bottomrule
\end{tabular}
\end{table}

\textbf{End-to-end speedup benchmark.} We incorporate the end-to-end efficiency metric including FLOPS, latency, and corresponding speedup into Table~\ref{table:accuracy_efficiency_benchmark}. \sys{} consistently outperforms all baselines by achieving an average speedup of $2.28\times$ while maintaining the highest generation quality. We further provide a detailed breakdown of end-to-end inference time on HunyuanVideo in Figure~\ref{fig:efficiency-breakdown-figure} to analyze the speedup. Each design point described in Sec~\ref{sec:methodology} contributes significantly to the speedup, with sparse attention delivering the most substantial improvement of $1.81\times$.

\begin{figure}[t]
    \centering
    \includegraphics[width=\columnwidth]{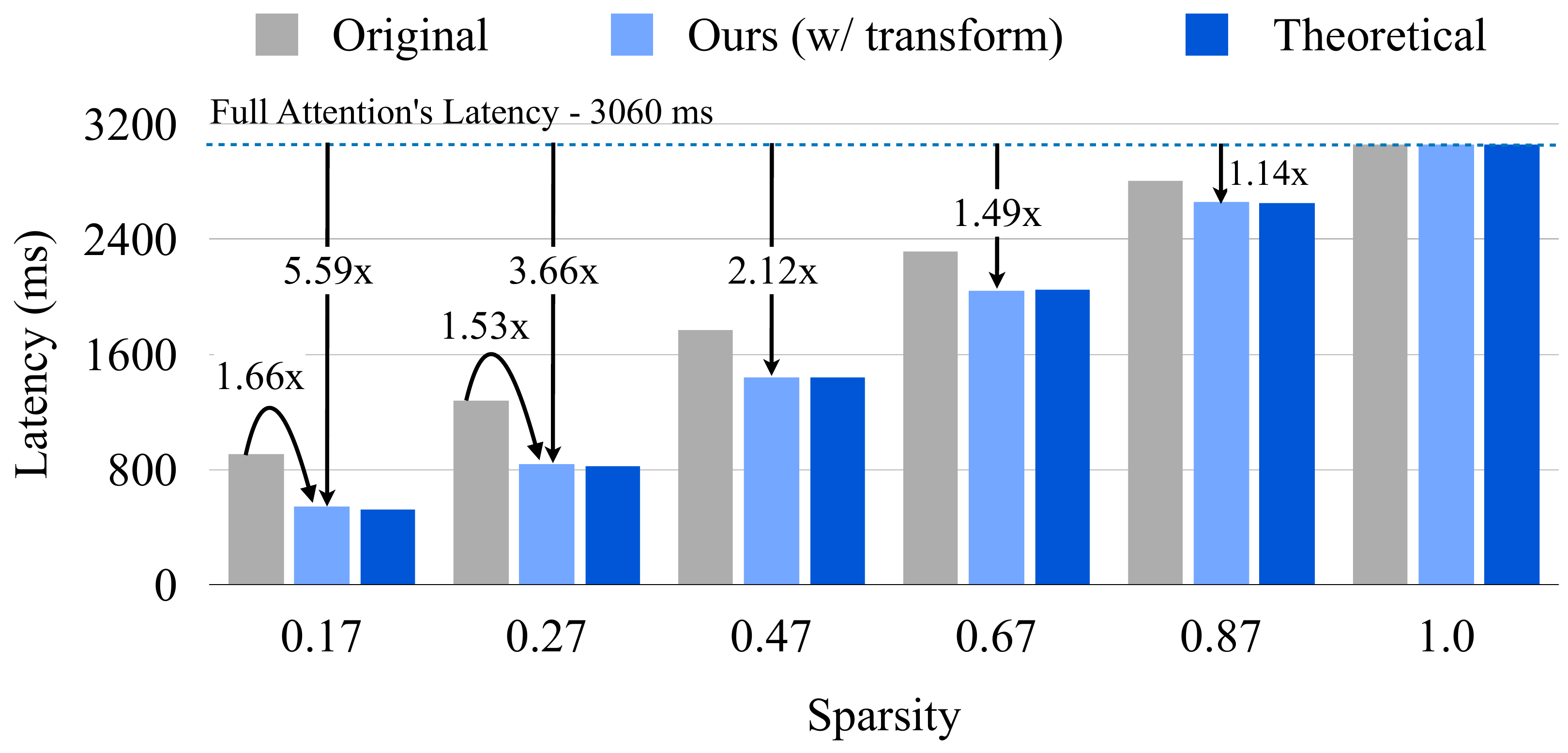} 
    \caption{Latency comparison of different implementations of sparse attention. Our hardware-efficient \reorder{} optimizes the sparsity pattern of \temporalhead{} for better contiguity, which is $1.7$× faster than naive sparse attention (named original), approaching the theoretical speedup.}
    \label{fig:kernel-efficiency-sparse-attention}
    \vspace{-5pt}
\end{figure}

\textbf{Kernel-level efficiency benchmark.}\label{subsec:kernel_level_efficiency} We benchmark individual kernel performance including QK-norm, RoPE, and block sparse attention with unit tests in Table~\ref{table:small-kernel-speedup-comparison}. Our customized QK-norm and RoPE achieve consistently better throughput across all frame numbers, with an average speedup of $7.4\times$ and $15.5\times$, respectively. For the sparse attention kernel, we compare the latency of our customized kernel with the theoretical speedup across different sparsity. As shown in Figure~\ref{fig:kernel-efficiency-sparse-attention}, our kernel achieves theoretical speedup, enabling practical benefit from sparse attention.

\begin{table}[t]
\centering
\caption{Sensitivity test on \onlinesample{} ratios. Profiling just $1$\% tokens achieves generation quality comparable to the oracle ($100$\%) while introducing only negligible overhead.}
\label{table:sensitivity-sampling}
\begin{tabular}{l|ccc}
\toprule
\textbf{Ratios} & \textbf{PSNR $\uparrow$} & \textbf{SSIM $\uparrow$} & \textbf{LPIPS $\downarrow$} \\
\midrule
\multicolumn{4}{c}{\textbf{CogVideoX-v1.5-I2V (720p, 10s, 80 frames)}} \\
\midrule
profiling 0.1\% & 30.791 & 0.941 & 0.0799 \\
profiling 1\% & 31.118 & 0.945 & 0.0757\\
profiling 5\% & 31.008 & 0.944 & 0.0764\\
profiling 100\% & 31.324 & 0.947 & 0.0744 \\
\bottomrule
\end{tabular}
\end{table}

\subsection{Sensitivity test}
\label{subsec:sensitivity-test}
In this section, we conduct a sensitivity analysis on the hyperparameter choices of \sys{}, including the \onlinesample{} ratios (Sec~\ref{subsec:sampling_based_pattern_selection}) and the sparsity ratios $c_s$ and $c_t$ (Sec~\ref{subsec:frame_token_rearrangement}). Our goal is to demonstrate the robustness of \sys{} across various efficiency-accuracy trade-offs.

\textbf{\Onlinesample{} ratios.} We evaluate the effectiveness of \onlinesample{} with different profiling ratios on CogVideoX-v1.5 using a random subset of VBench in Table~\ref{table:sensitivity-sampling}. In our experiments, we choose to profile only 1\% of the input rows, which offers a comparable generation quality comparable to the oracle profile (100\% profiled) with negligible overhead.


\textbf{Generation quality over different sparsity ratios.} As discussed in Sec~\ref{sec:sparse-theoretical-speedup}, different sparsity ratio of the \spatialhead{} and \temporalhead{} can be set by choosing different $c_s$ and $c_t$, therefore reaching different trade-offs between efficiency and accuracy. We evaluate the LPIPS of HunyuanVideo over a random subset of VBench with different sparsity ratios. As shown in Table~\ref{table:sensitivity-sparsity-ratios}, \sys{} consistently achieves decent generation quality across various sparsity ratios. E.g., even with a sparsity of $13$\%, \sys{} still achieves $0.154$ LPIPS. We leave the adaptive sparsity control for future work.

\subsection{Ablation study}
\label{subsec:ablation}
We conduct the ablation study to evaluate the effectiveness of the proposed hardware-efficient \reorder{} (as discussed in Sec~\ref{subsec:frame_token_rearrangement}). Specifically, we profile the latency of the sparse attention kernel with and without the transformation under the HunyuanVideo configuration. As shown in Figure~\ref{fig:kernel-efficiency-sparse-attention}, the sparse attention with \reorder{} closely approaches the theoretical speedup, whereas the original implementation without \reorder{} falls short. For example, at a sparsity level of $10$\%, our method achieves an additional $1.7\times$ speedup compared to the original approach, achieving a $3.63\times$ improvement.

\begin{table}[t]
\small
\centering
\caption{Video quality of HunyuanVideo on a subset of VBench when varying sparsity ratios. LPIPS decreases as the sparse ratio increases, achieving trade-offs between efficiency and accuracy.}
\label{table:sensitivity-sparsity-ratios}
\begin{tabular}{c|cccccc}
\toprule
Sparsity$\downarrow$ & 0.13 & 0.18 & 0.35 & 0.43 & 0.52 \\
\midrule
LPIPS$\downarrow$ & 0.154 & 0.135 & 0.141 & 0.129 & 0.116 \\
\bottomrule
\end{tabular}
\vspace{-5pt}
\end{table}


\section{Conclusion}
We accelerate video diffusion transformers by exploiting sparse attention. We reveal that attention heads have inherent sparsity patterns and we classify them into spatial head and temporal head. We proposed Sparse VideoGen (SVG), a training-free method to utilize these sparsity patterns for end-to-end efficiency boosts, including an efficient online profiling algorithm and an efficient inference system. On representative open video diffusion transformers (CogVideoX-v1.5 and HunyuanVideo), SVG demonstrates prominent end-to-end speedup without losing quality.

\section*{Impact Statement}
This paper presents work that aims to advance the field of Machine Learning. There are many potential societal consequences of our work, none of which we feel must be specifically highlighted here.

\section*{Acknowledgment}
We thank Guangxuan Xiao and Zihao Ye for the visualizations and kernel designs.

\nocite{langley00}

\bibliography{example_paper}
\bibliographystyle{icml2025}

\newpage
\appendix
\onecolumn




    

\section{A full version of related work}

\subsection{Efficient Diffusion Models}\label{subsec:efficient_diffusion}
Diffusion Models function primarily as denoising models that are trained to estimate the gradient of the data distribution \citep{song2019generative,zhang2023tale}. Although these models are capable of generating samples with high quality and diversity, they are known as inefficient. To enhance the efficiency of diffusion models, researchers often focus on three primary approaches: (1) decreasing the number of denoising steps, (2) reducing the model size, and (3) optimizing system implementation for greater efficiency.

\paragraph{Decreasing the denoising steps.} 
The main diffusion models rely on stochastic differential equations (SDEs) that learn to estimate the gradient of the data distribution through Langevin dynamics \citep{ho2020denoising, meng2022sdedit}. Consequently, these models generally require numerous sampling steps (\textit{, e.g.,} 1,000). To improve sample efficiency, DDIM \citep{song2020denoising} approximates SDE-based diffusion models within an ordinary differential equation (ODE) framework. Expanding on this concept, DPM \citep{lu2022dpm}, DPM++ \citep{lu2022dpm++}, and Rectified Flows \citep{liu2022flow, liu2023instaflow} enhance ODE paths and solvers to further reduce the number of denoising steps. Furthermore, Consistency Models \citep{song2023consistency, luo2023latent} integrate the ODE solver into training using a consistency loss, allowing diffusion models to replicate several denoising operations with fewer iterations. In addition, approaches grounded in distillation \citep{yin2024improved,yin2024one} represent another pivotal strategy. This involves employing a simplified, few-step denoising model to distill a more complex, multi-step denoising model, thereby improving overall efficiency.

Nevertheless, all these approaches necessitate either re-training or fine-tuning the complete models on image or video datasets. For video generation models, this is largely impractical due to the significant computational expense involved, which is prohibitive for the majority of users. In this work, our primary focus is on a method to enhance generation speed that requires no additional training.

\paragraph{Diffusion Model Compreesion}
A common approach to enhancing the efficiency of diffusion models involves compressing their weights through quantization. Q-Diffusion~\citep{li2023q} introduced a W8A8 strategy, implementing quantization in these models. Building on this foundation, ViDiT-Q~\citep{zhao2024vidit} proposed a timestep-aware dynamic quantization method that effectively reduces the bit-width to W4A8. Furthermore, SVDQuant~\citep{li2024svdquant} introduced a cost-effective branch designed to address outlier problems in both activations and weights, thus positioning W4A4 as a feasible solution for diffusion models. SageAttention~\citep{zhang2025sageattention} advanced the field by quantizing the attention module to INT8 precision via a smoothing technique. SageAttention V2~\citep{zhang2024sageattention2} extended these efforts by pushing the precision boundaries to INT4 and FP8. Another common approach is to design efficient diffusion model architectures \cite{xie2024sana,cai2024condition,chen2025pixart} and high-compression autoencoders \cite{chen2024deep} to boost efficiency. Our Sparse VideoGen is orthogonal to these techniques and can utilize them as supplementary methods to enhance efficiency.

\paragraph{Efficient System Implementation}
In addition to enhancing the efficiency of diffusion models by either retraining the model to decrease the number of denoising steps or compressing the model size, efficiency improvements can also be achieved at the system level. For instance, strategies such as dynamic batching are employed in StreamDiffusion~\citep{kodaira2023streamdiffusion} and StreamV2V~\citep{liang2024looking} to effectively manage streaming inputs in diffusion models, thereby achieving substantial throughput enhancements. Other approaches include: DeepCache~\citep{ma2024deepcache}, which leverages feature caching to modify the UNet Diffusion; $\Delta-DiT$~\citep{chen2024delta}, which implements this mechanism by caching residuals between attention layers in DiT to circumvent redundant computations; and PAB~\citep{zhao2024pab}, which caches and broadcasts intermediary features at distinct timestep intervals. FasterCache~\citep{lv2024fastercache} identifies significant redundancy in CFG and enhances the reuse of both conditional and unconditional outputs. Meanwhile, TeaCache~\cite{liu2024timestep} recognizes that the similarity in model inputs can be used to forecast output similarity, suggesting an improved machine strategy to amplify speed gains.

Despite these advanced methodologies, they often result in the generated output diverging significantly from the original, as indicated by a PSNR falling below 22. In contrast, our method consistently achieves a PSNR exceeding 30, thus ensuring substantially superior output quality compared to these previously mentioned strategies.

\subsection{Efficient Attention}

\paragraph{Sparse Attention in LLM} Recent studies on sparse attention in language models have identified patterns that reduce computational costs by targeting specific token subsets. StreamingLLM \cite{xiao2023efficient} and LM-Infinite \cite{han2023lm} reveal concentration on initial and local tokens, highlighting temporal locality in decoding. H2O \cite{zhang2023h2o} and Scissorhands \cite{liu2024scissorhands} note attention focuses mainly on a few dominant tokens. TidalDecode \cite{yang2024tidaldecode} shows cross-layer attention pattern correlations, aiding in attention sparsity. DuoAttention \cite{xiao2024duoattention} and MInference \cite{jiang2024minference} find distinct sparse patterns among attention heads, with varying focus on key tokens and context. MMInference ~\cite{li2025mminference} speedup the vision language models through modality-aware permutation. SpargeAttention \cite{zhang2025spargeattn} and XAttention \cite{xu2025xattention} propose general sparsity identification algorithms that can be applied to all forms of models. Despite their success in LLMs or VLMs, these mechanisms are constrained to token-level sparsity and miss the redundancy unique to video data. 

\paragraph{Linear and Low-bit Attention} Significant advancements have been achieved in enhancing attention efficiency, notably through linear attention \cite{cai2023efficientvit,xie2024sana} and low-bit attention techniques \cite{zhang2025sageattention,zhang2024sageattention2}. Linear attention models, including Linformer \cite{wang2020linformer}, Performer \cite{choromanski2020rethinking}, MetaFormer \cite{yu2022metaformer}, and LinearAttention \cite{katharopoulos2020transformers}, reduce the quadratic complexity of traditional attention to linear. Low-bit attention approaches decrease computational demands by utilizing lower precision, with SageAttention \cite{zhang2025sageattention} employing INT8 precision to enhance efficiency without notable performance loss.
 
Sparse VideoGen, as a \textbf{sparse attention} method, is \textbf{orthogonal} to both linear attention and low-bit attention techniques. 
Moreover, it can be integrated with low-bit attention methods, such as FP8 attention, to further enhance computational efficiency. 

\section{Visualization of the generated videos}\label{appendix: visualization}
We provide visualization comparison between Dense Attention and Sparse VideoGen on HunyuanVideo and Wan 2.1. We conduct both Text-to-Video generation and Image-to-Video generation under 720p resolution. Results demonstrates that Sparse VideoGen can preserve high pixel-level fidelity, achieving similar generation quality compared with the dense attention.

\begin{figure}
    \centering
    \begin{minipage}{0.15\linewidth}
        \centering
        \small 
        Dense Attention
    \end{minipage}
    \hfill
    \begin{minipage}{0.84\linewidth}
        \includegraphics[width=\linewidth]{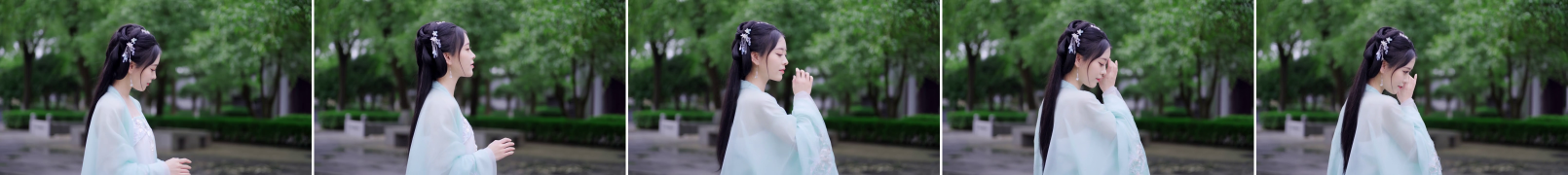}
    \end{minipage}

    \begin{minipage}{0.15\linewidth}
        \centering
        \small 
        Sparse VideoGen
    \end{minipage}
    \hfill
    \begin{minipage}{0.84\linewidth}
        \includegraphics[width=\linewidth]{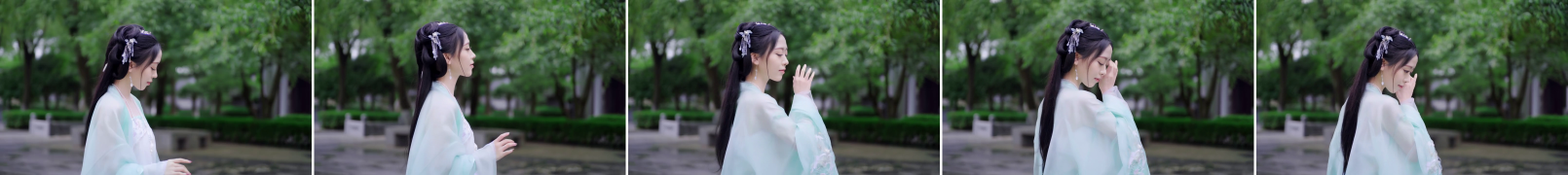}
    \end{minipage}

    \vspace{5pt}

    \begin{minipage}{0.15\linewidth}
        \centering
        \small 
        Dense Attention
    \end{minipage}
    \hfill
    \begin{minipage}{0.84\linewidth}
        \includegraphics[width=\linewidth]{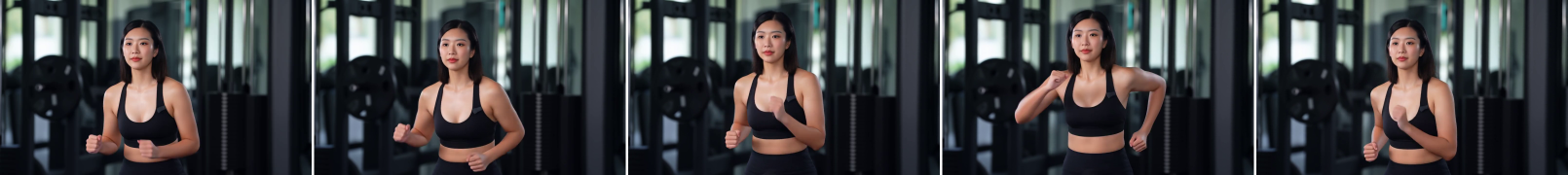}
    \end{minipage}

    \begin{minipage}{0.15\linewidth}
        \centering
        \small 
        Sparse VideoGen
    \end{minipage}
    \hfill
    \begin{minipage}{0.84\linewidth}
        \includegraphics[width=\linewidth]{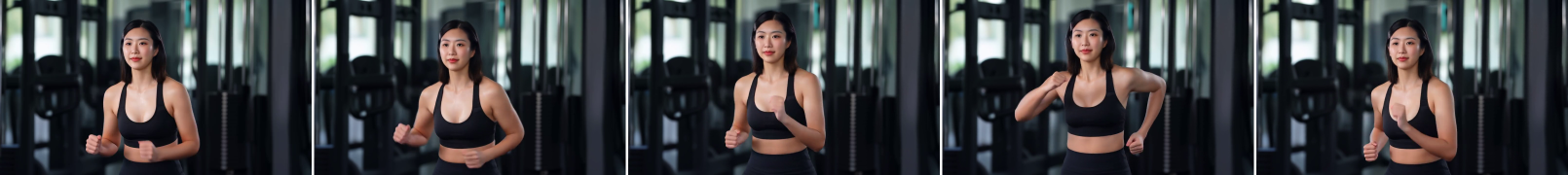}
    \end{minipage}
    
    \vspace{5pt}

    \begin{minipage}{0.15\linewidth}
        \centering
        \small 
        Dense Attention
    \end{minipage}
    \hfill
    \begin{minipage}{0.84\linewidth}
        \includegraphics[width=\linewidth]{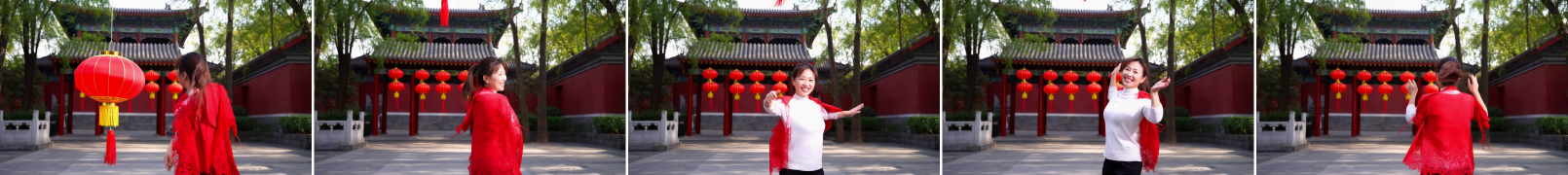}
    \end{minipage}

    \begin{minipage}{0.15\linewidth}
        \centering
        \small 
        Sparse VideoGen
    \end{minipage}
    \hfill
    \begin{minipage}{0.84\linewidth}
        \includegraphics[width=\linewidth]{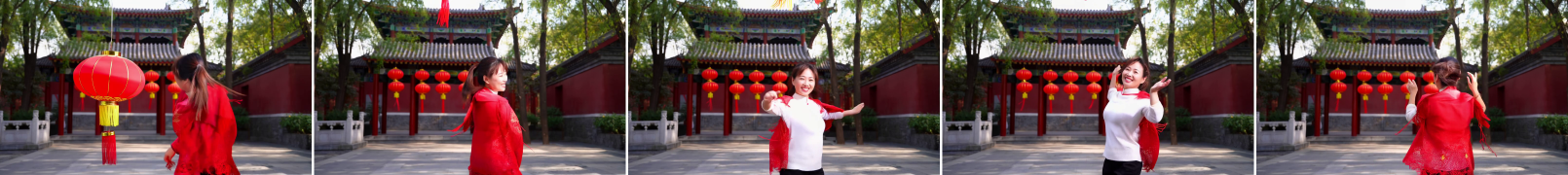}
    \end{minipage}
        
    \vspace{5pt}

    \begin{minipage}{0.15\linewidth}
        \centering
        \small 
        Dense Attention
    \end{minipage}
    \hfill
    \begin{minipage}{0.84\linewidth}
        \includegraphics[width=\linewidth]{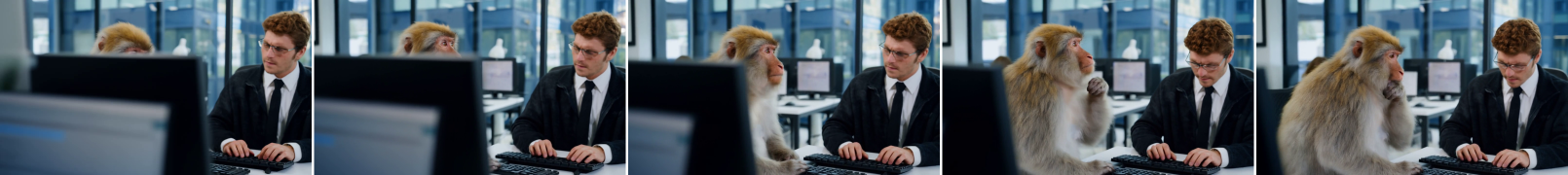}
    \end{minipage}

    \begin{minipage}{0.15\linewidth}
        \centering
        \small 
        Sparse VideoGen
    \end{minipage}
    \hfill
    \begin{minipage}{0.84\linewidth}
        \includegraphics[width=\linewidth]{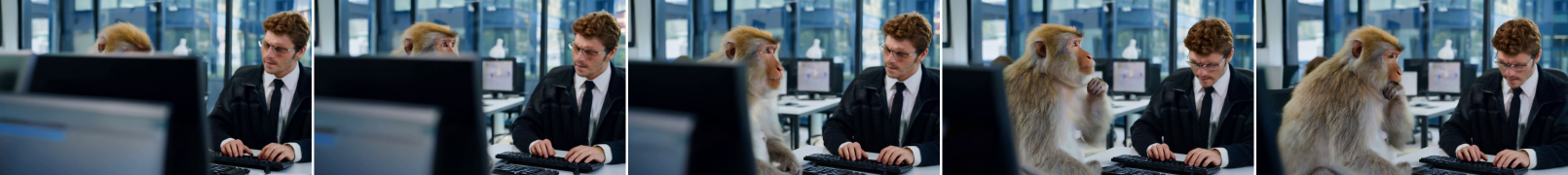}
    \end{minipage}
            
    \vspace{5pt}

    \begin{minipage}{0.15\linewidth}
        \centering
        \small 
        Dense Attention
    \end{minipage}
    \hfill
    \begin{minipage}{0.84\linewidth}
        \includegraphics[width=\linewidth]{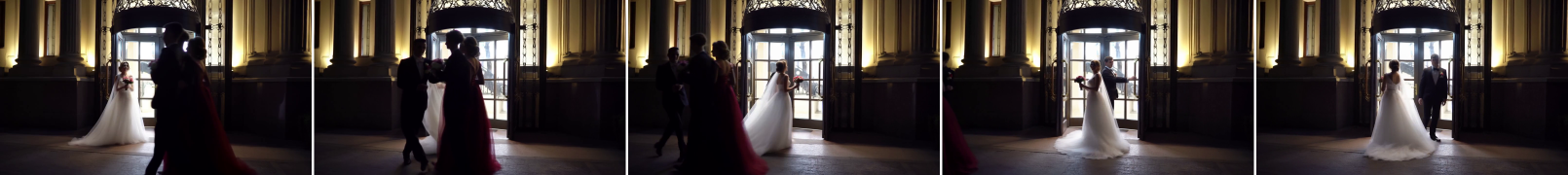}
    \end{minipage}

    \begin{minipage}{0.15\linewidth}
        \centering
        \small 
        Sparse VideoGen
    \end{minipage}
    \hfill
    \begin{minipage}{0.84\linewidth}
        \includegraphics[width=\linewidth]{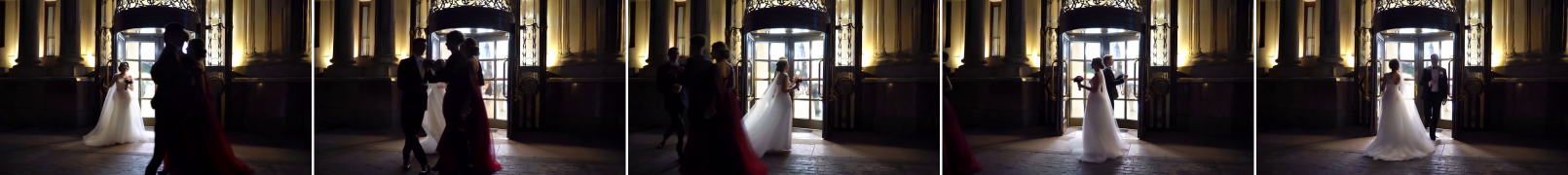}
    \end{minipage}
    
    \caption{Comparion of Dense Attention and Sparse VideoGen on HunyuanVideo Text-to-Video generation.}
    \label{fig:Hunyuan_T2V_Visualization}
\end{figure}

\begin{figure}
    \centering
    \begin{minipage}{0.15\linewidth}
        \centering
        \small 
        Dense Attention
    \end{minipage}
    \hfill
    \begin{minipage}{0.84\linewidth}
        \includegraphics[width=\linewidth]{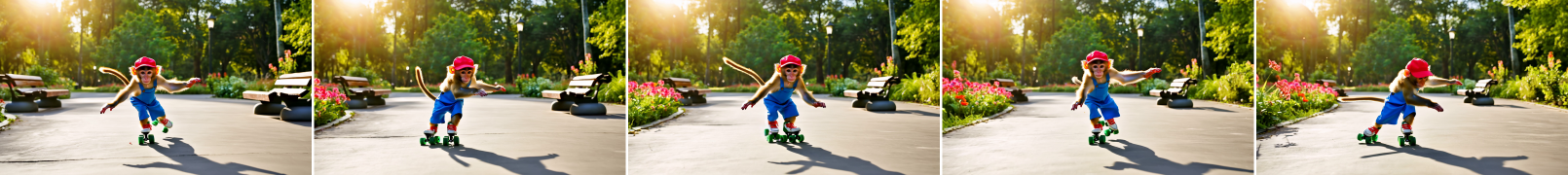}
    \end{minipage}

    \begin{minipage}{0.15\linewidth}
        \centering
        \small 
        Sparse VideoGen
    \end{minipage}
    \hfill
    \begin{minipage}{0.84\linewidth}
        \includegraphics[width=\linewidth]{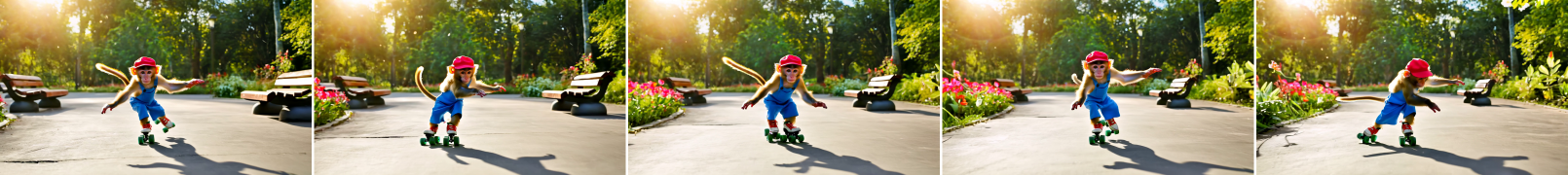}
    \end{minipage}

    \vspace{5pt}
    
    \begin{minipage}{0.15\linewidth}
        \centering
        \small 
        Dense Attention
    \end{minipage}
    \hfill
    \begin{minipage}{0.84\linewidth}
        \includegraphics[width=\linewidth]{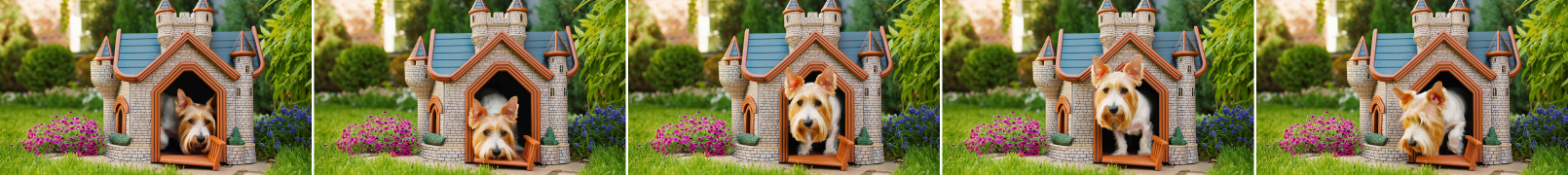}
    \end{minipage}

    \begin{minipage}{0.15\linewidth}
        \centering
        \small 
        Sparse VideoGen
    \end{minipage}
    \hfill
    \begin{minipage}{0.84\linewidth}
        \includegraphics[width=\linewidth]{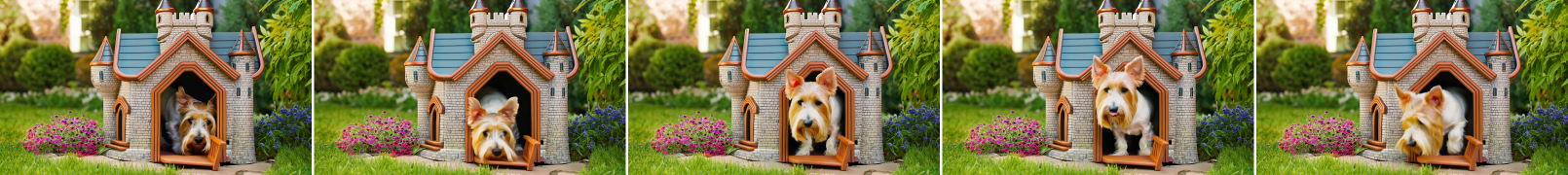}
    \end{minipage}

    \vspace{5pt}
    
    \begin{minipage}{0.15\linewidth}
        \centering
        \small 
        Dense Attention
    \end{minipage}
    \hfill
    \begin{minipage}{0.84\linewidth}
        \includegraphics[width=\linewidth]{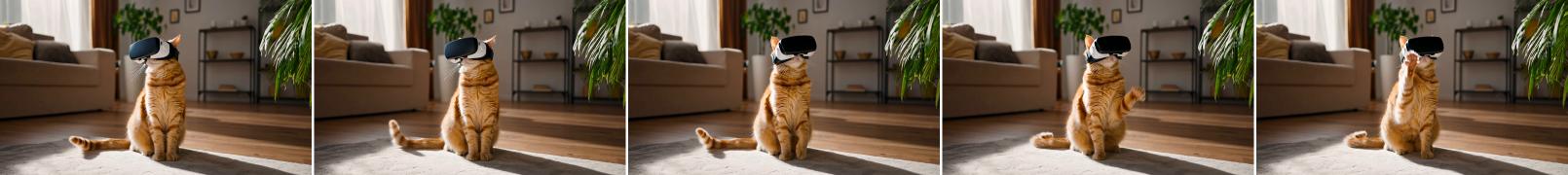}
    \end{minipage}

    \begin{minipage}{0.15\linewidth}
        \centering
        \small 
        Sparse VideoGen
    \end{minipage}
    \hfill
    \begin{minipage}{0.84\linewidth}
        \includegraphics[width=\linewidth]{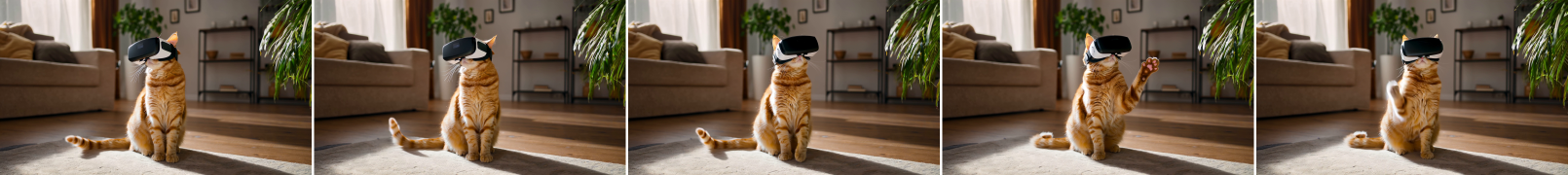}
    \end{minipage}

    \vspace{5pt}
    
    \begin{minipage}{0.15\linewidth}
        \centering
        \small 
        Dense Attention
    \end{minipage}
    \hfill
    \begin{minipage}{0.84\linewidth}
        \includegraphics[width=\linewidth]{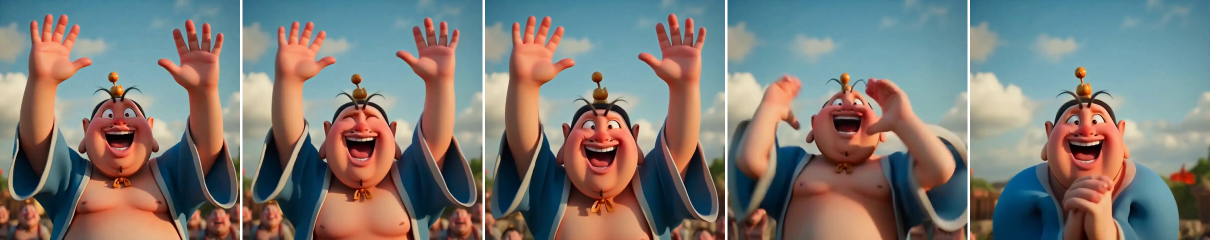}
    \end{minipage}

    \begin{minipage}{0.15\linewidth}
        \centering
        \small 
        Sparse VideoGen
    \end{minipage}
    \hfill
    \begin{minipage}{0.84\linewidth}
        \includegraphics[width=\linewidth]{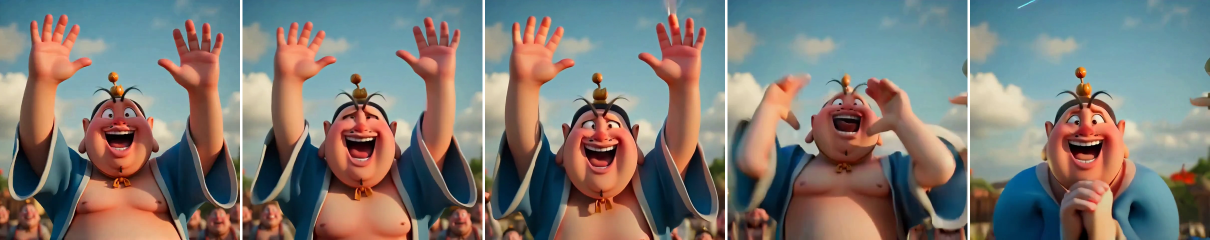}
    \end{minipage}

    \caption{Comparion of Dense Attention and Sparse VideoGen on Wan 2.1 Text-to-Video generation.}
    \label{fig:Wan_T2V_Visualization}
\end{figure}

\begin{figure}
    \centering

    \begin{minipage}{0.48\linewidth}
        \centering
        \text{Dense Attention}\\[0.5em] 
        \includegraphics[width=\textwidth]{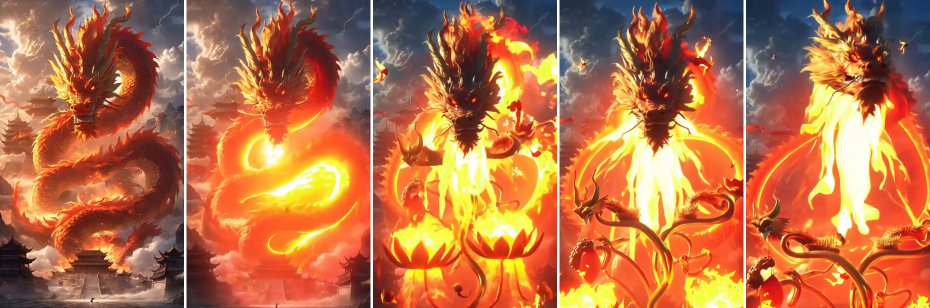}
    \end{minipage}
    \hfill
    \begin{minipage}{0.48\linewidth}
        \centering
        \text{Sparse VideoGen}\\[0.5em] 
        \includegraphics[width=\textwidth]{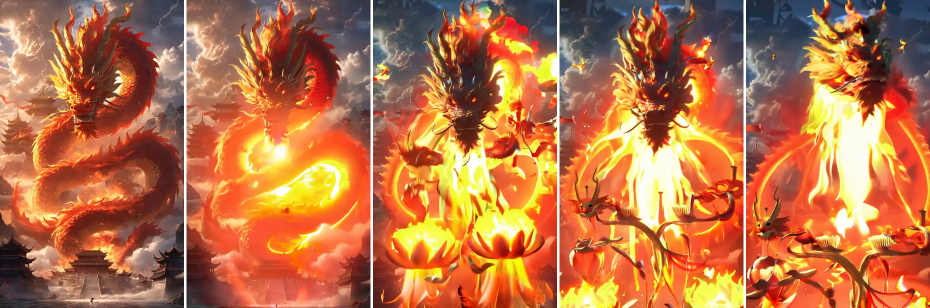}
    \end{minipage}
    
    \vspace{10pt}

    \begin{minipage}{0.48\linewidth}
        \centering
        \text{Dense Attention}\\[0.5em] 
        \includegraphics[width=\textwidth]{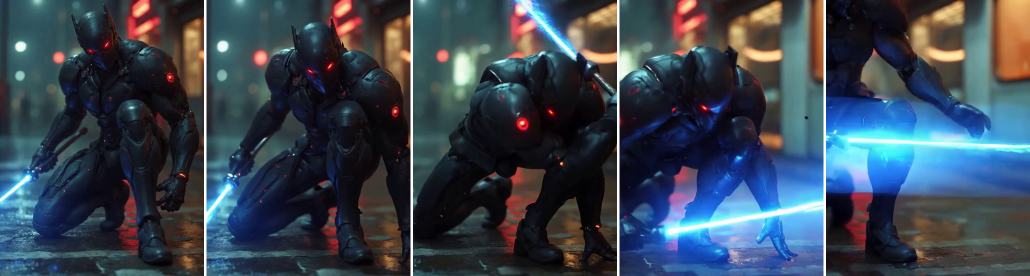}
    \end{minipage}
    \hfill
    \begin{minipage}{0.48\linewidth}
        \centering
        \text{Sparse VideoGen}\\[0.5em] 
        \includegraphics[width=\textwidth]{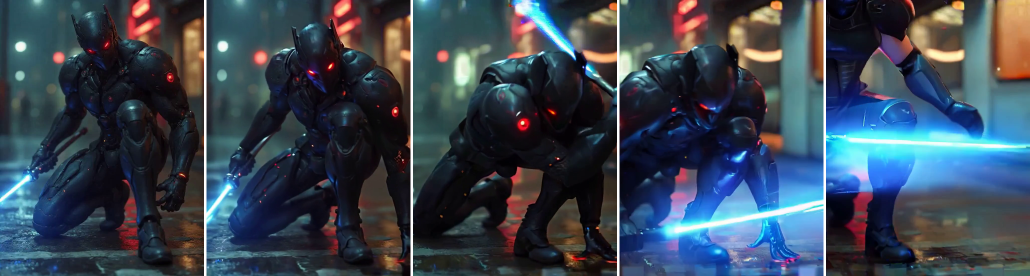}
    \end{minipage}
    
    \vspace{10pt}
    
    \begin{minipage}{0.48\linewidth}
        \centering
        \text{Dense Attention}\\[0.5em] 
        \includegraphics[width=\textwidth]{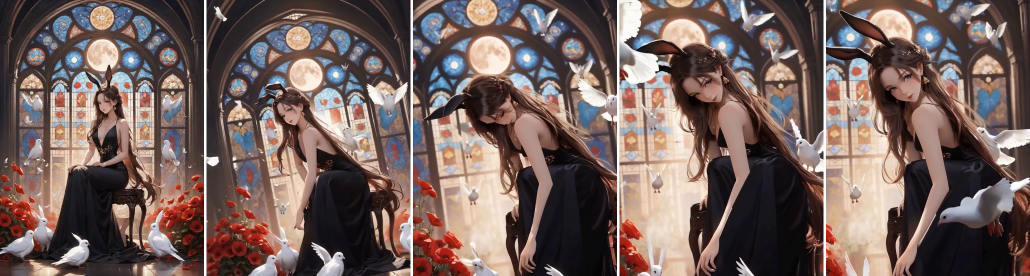}
    \end{minipage}
    \hfill
    \begin{minipage}{0.48\linewidth}
        \centering
        \text{Sparse VideoGen}\\[0.5em] 
        \includegraphics[width=\textwidth]{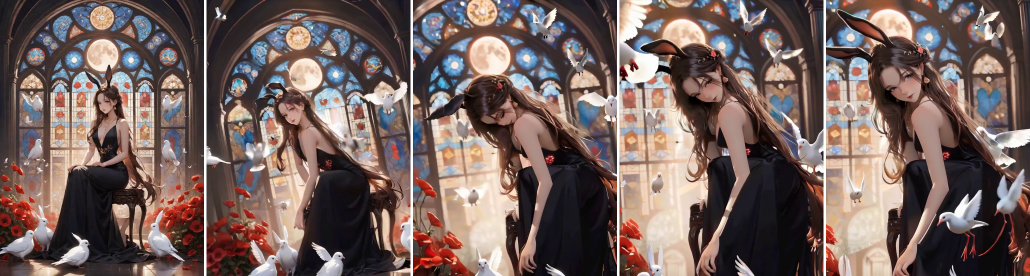}
    \end{minipage}
    
    \vspace{10pt}

    \begin{minipage}{0.48\linewidth}
        \centering
        \text{Dense Attention}\\[0.5em] 
        \includegraphics[width=\textwidth]{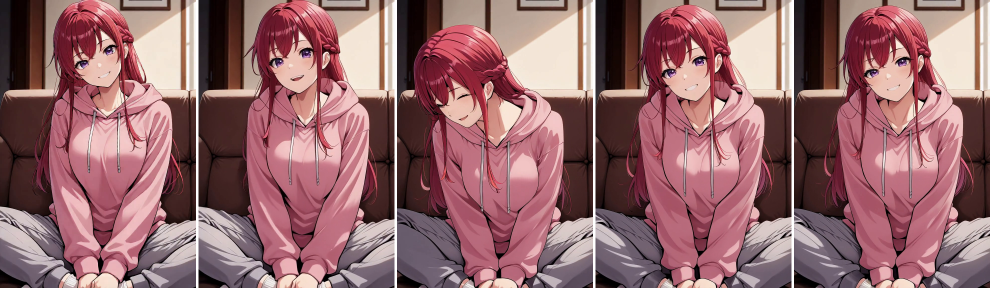}
    \end{minipage}
    \hfill
    \begin{minipage}{0.48\linewidth}
        \centering
        \text{Sparse VideoGen}\\[0.5em] 
        \includegraphics[width=\textwidth]{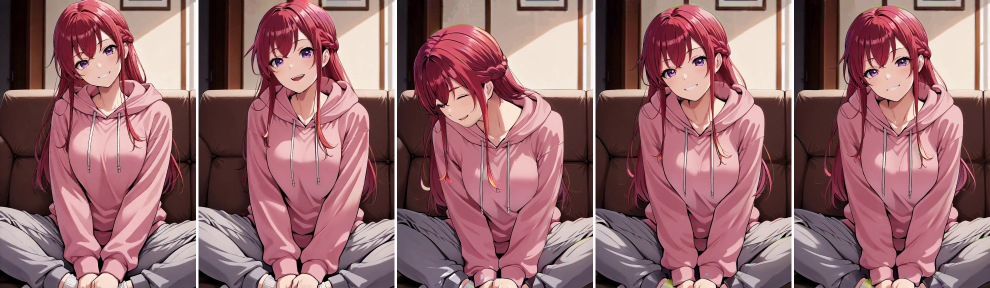}
    \end{minipage}
    
    \vspace{10pt}

    \begin{minipage}{0.48\linewidth}
        \centering
        \text{Dense Attention}\\[0.5em] 
        \includegraphics[width=\textwidth]{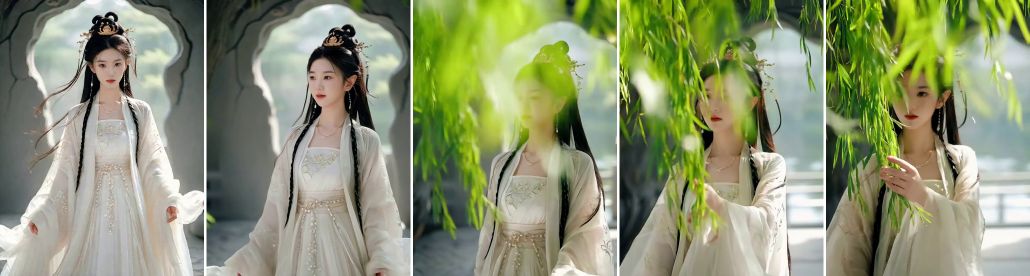}
    \end{minipage}
    \hfill
    \begin{minipage}{0.48\linewidth}
        \centering
        \text{Sparse VideoGen}\\[0.5em] 
        \includegraphics[width=\textwidth]{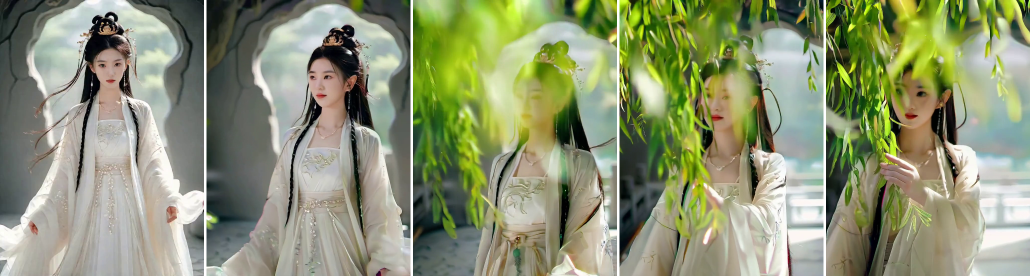}
    \end{minipage}
    
    \vspace{20pt}

    \begin{minipage}{0.15\linewidth}
        \centering
        \small 
        Dense Attention
    \end{minipage}
    \hfill
    \begin{minipage}{0.84\linewidth}
        \includegraphics[width=\linewidth]{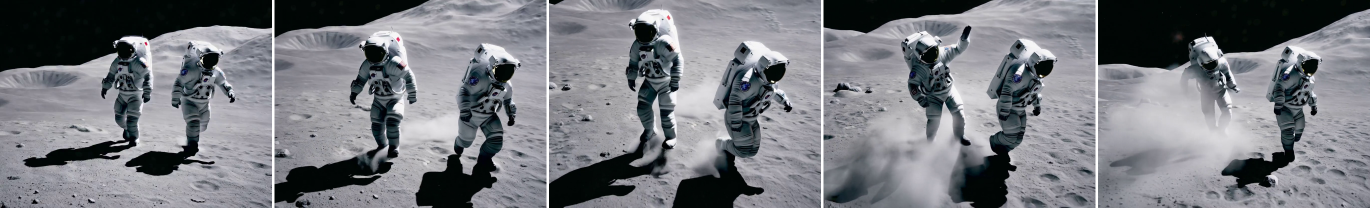}
    \end{minipage}

    \begin{minipage}{0.15\linewidth}
        \centering
        \small 
        Sparse VideoGen
    \end{minipage}
    \hfill
    \begin{minipage}{0.84\linewidth}
        \includegraphics[width=\linewidth]{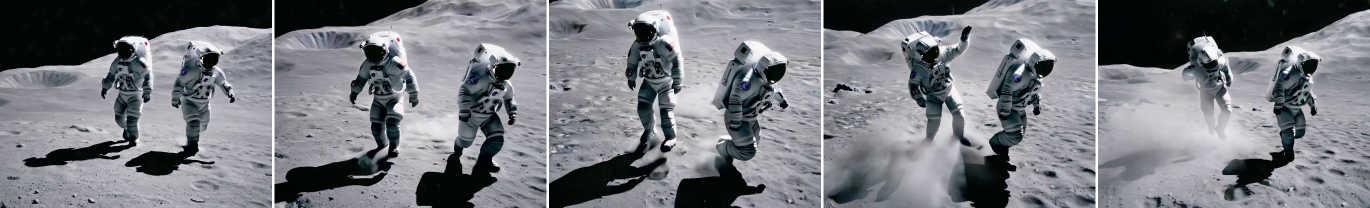}
    \end{minipage}

    \caption{Comparison of Dense Attention and Sparse VideoGen on Wan 2.1 Image-to-Video generation.}
    \label{fig:Wan_I2V_Visualization}
\end{figure}


\end{document}